\documentclass[10pt]{wlscirep}
\usepackage[utf8]{inputenc}
\usepackage[T1]{fontenc}
\usepackage{amsmath}
\usepackage{bm}
\usepackage{multirow}
\usepackage[draft]{pgf}
\usepackage{hyperref}
\usepackage{makecell}
\usepackage{float}
\usepackage{lineno}
\newcommand\setcurrentname[1]{\def\@currentlabelname{#1}}
\DeclareMathAlphabet{\mathcal}{OMS}{cmsy}{m}{n}

\makeatletter
\def\thickhline{%
  \noalign{\ifnum0=`}\fi\hrule \@height \thickarrayrulewidth \futurelet
   \reserved@a\@xthickhline}
\def\@xthickhline{\ifx\reserved@a\thickhline
    \vskip\doublerulesep
    \vskip-\thickarrayrulewidth
    \fi
    \ifnum0=`{\fi}}
\makeatother
\newlength{\thickarrayrulewidth}
\newcommand{\pjremark}[1]{#1}

\title{An Empirical Study of Large-Scale Data-Driven Full Waveform Inversion}

\author[1,2, *]{Peng Jin}
\author[1]{Yinan Feng}
\author[1]{Shihang Feng}
\author[1]{Hanchen Wang}
\author[3]{Yinpeng Chen}
\author[4]{\\Benjamin Consolvo}
\author[5]{Zicheng Liu}
\author[6,*]{Youzuo Lin}
\affil[1]{Earth and Environmental Sciences Division, Los Alamos National Laboratory}
\affil[2]{College of Information Sciences and Technology, The Pennsylvania State University}
\affil[3]{Google Research}
\affil[4]{Intel}
\affil[5]{Microsoft}
\affil[6]{School of Data Science and Society, The University of North Carolina at Chapel Hill}
\affil[*]{Corresponding Authors: pqj5125@psu.edu, yzlin@unc.edu}

\begin{abstract}
This paper investigates the impact of big data on deep learning models to help solve the full waveform inversion (FWI) problem. While it is well known that big data can boost the performance of deep learning models in many tasks, its effectiveness has not been validated for FWI. To address this gap, we present an empirical study that investigates how deep learning models in FWI behave when trained on \textsc{OpenFWI}, a collection of large-scale, multi-structural, synthetic datasets published recently. In particular, we train and evaluate the FWI models on a combination of 10 2D subsets in \textsc{OpenFWI} that contain 470K pairs of seismic data and velocity maps in total. \pjremark{Our experiments demonstrate that training on the combined dataset yields an average improvement of 13.03\% in MAE, 7.19\% in MSE and 1.87\% in SSIM compared to each split dataset, and an average improvement of 28.60\%, 21.55\% and 8.22\% in the leave-one-out generalization test. We further demonstrate that model capacity needs to scale in accordance with data size for optimal improvement, where our largest model yields an average improvement of 20.06\%, 13.39\% and 0.72\% compared to the smallest one.}

\end{abstract}
\begin{document}

\flushbottom
\maketitle
% \thispagestyle{empty}

% \noindent Please note: Abbreviations should be introduced at the first mention in the main text – no abbreviations lists. Suggested structure of main text (not enforced) is provided below.

\section*{Introduction}
% Big data intro
The recent advancements of deep learning in natural language processing and computer vision have proven that big data is one of the key ingredients for obtaining good performance~\cite{brown2020language, sun2017revisiting, kolesnikov2020big, schuhmannlaion}. Similarly, in the context of science, deep learning models such as AlphaFold~\cite{jumper2021highly} have achieved significant breakthroughs with the help of large-scale datasets. However, unlike these tasks, large-scale public datasets are not always available for many other scientific problems due to issues such as high data acquisition costs, labeling costs, intellectual property concerns, or security concerns. Due to limited dataset sizes and variation, deep learning models in scientific applications are often limited in their ability to generalize well to out-of-sample datasets.

\begin{figure}
    \centering
    \includegraphics[width=\textwidth]{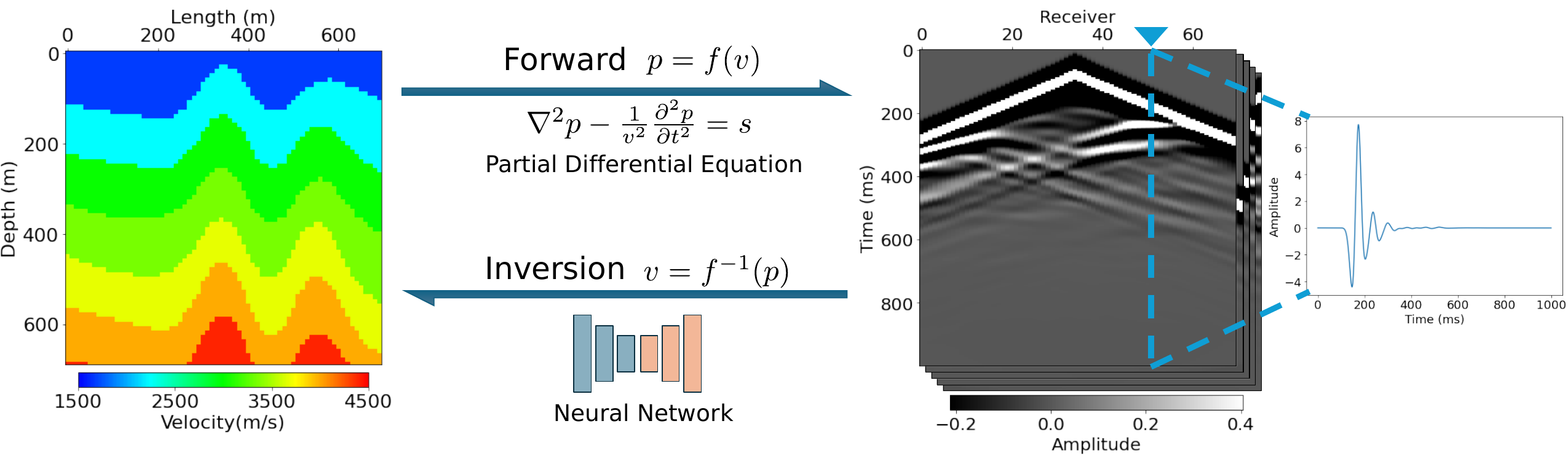}
    \caption{\textbf{Schematic illustration of data-driven FWI and forward modeling}. The forward modeling process computes the simulated seismic data from velocity maps, governed by a partial differential equation. Neural networks are employed in data-driven FWI methods to reconstruct velocity maps from seismic measurements.}
    \label{fig:fwi}
\end{figure}

% - FWI intro, OpenFWI
Full waveform inversion (FWI) is a technique used to image the subsurface that has the potential to benefit from deep learning and large training datasets. Specifically, FWI aims to reconstruct subsurface velocity maps $v$ from seismic measurements $p$ as depicted in Figure~\ref{fig:fwi}. Conventional FWI methods~\cite{tarantola1984inversion, pratt1998gauss, plessix2006review, virieux2009overview, fichtner2010full, zhang2012wave,ma2012image, zhang2013double,feng2019transmission+,feng2021mpi, lin2014acoustic, lin2015quantifying, hu2009simultaneous, guitton2012blocky,chen2020multiscale} leverage the forward operator $f$ governed by a partial differential equation (PDE) and perform iterative optimization per sample, which is computationally expensive and yields poor scalability. To mitigate this issue, deep learning techniques have been recently introduced to FWI and achieved promising performance~\cite{wang2020velocity, liu2021deep, araya2018deep, yang2019deep, ren2021building, geng2022deep}. A good summary of deep learning techniques for solving FWI problems can be found in Lin et al.~\cite{lin2023physics}. In this paper, we follow the previous studies~\cite{wu2019inversionnet,zhang2020data,lin2023physics} and refer network-based FWI methods as data-driven methods. Inspired by the image-to-image translation task in computer vision, these data-driven methods directly learn an inverse mapping $f^{-1}$ from seismic data directly to velocity maps. Nevertheless, due to the issue of lacking large-scale public datasets, the models in prior works were all developed on relatively small datasets (i.e. 130 to 67K data pairs)~\cite{araya2017automated,wu2019inversionnet,zhang2020data}. Thus, the question remains open: \textit{does full waveform inversion benefit from big data?} Thankfully, the recently published large-scale datasets \textsc{OpenFWI}~\cite{deng2022openfwi} provide us an opportunity to start to answer this question. % does full waveform inversion benefit from big data? do deep learning models in FWI benefit from big data?

% - Project overview
% - Contribution
In this paper, we present an empirical study that attempts to answer the question of whether FWI benefits from large-scale and multi-structural training datasets from three perspectives: model performance, the relationship between the model size and data size, and model generalization. \textsc{OpenFWI} is a collection of large-scale, multi-structural datasets that cover different domain interests, including interfaces, geological faults, and field data. We employ 10 2D synthetic datasets from \textsc{OpenFWI}, and 408K/62K pairs of seismic data and velocity maps are used to train and evaluate the deep learning models, respectively. We adopt one of the \textsc{OpenFWI} benchmark models InversionNet~\cite{wu2019inversionnet} to serve as the baseline, and we compare the inversion results of the baselines trained on relatively small-scale individual datasets and the models trained on large-scale datasets that are composed of multiple datasets. We name the latter models \textit{BigFWI}. Our findings are summarized as follows: 
% We employ the InversionNet~\cite{wu2019inversionnet} as the shared network architecture across all experiments.
% and we hope this study can serve as a guideline for future data-driven full waveform inversion research.
\begin{itemize}[topsep=0pt] %,noitemsep]
    \itemsep -0.5em
    % FWI needs big data
    \item \textbf{Big data can boost the performance of deep learning models in FWI.} BigFWI outperforms the baselines on almost every dataset in terms of all the evaluation metrics. 
    \item \textbf{Larger data requires larger models.} When more training samples are introduced, larger model architectures are required in BigFWI to achieve further improvement compared to the baselines. 
    % When half of each dataset is used for training (i.e. 204K data pairs in total), BigFWI outperforms the baselines on almost every dataset by an average improvement of 13.03\% in mean absolute error (MAE), 7.19\% in root mean squared error (RMSE) and 1.87\% in structural similarity (SSIM). 
    % FWI needs big model for big data
    \item \textbf{Big data can improve the generalization of deep learning models in FWI.} Given a dataset that is unseen during training, BigFWI yields better performance than any baselines trained on single datasets. 
    % When more training data is included, the benefit of combination diminishes. However, we can retain the improvement by increasing the model size. BigFWI with a larger model size outperforms the baseline models by an average improvement of 18.66\% in MAE, 13.62\% in RMSE, and 2.17\% in SSIM. 
\end{itemize}

\vspace{-5pt}

\section*{Methods}
\pjremark{In this section, we first present the preliminaries of full waveform inversion and then describe the network architecture of our BigFWI and the loss function for training.}

\subsection*{Full Waveform Inversion}
% Equation & Data-driven FWI
Figure~\ref{fig:fwi} provides an illustration of 2D data-driven FWI and forward modeling. The governing equation of the acoustic wave forward modeling in an isotropic medium with a constant density can be described as follows:
\begin{equation} \label{eq:1}
  \nabla^2 p(\bm{r}, t) - \frac{1}{v(\bm{r})^2} \frac{\partial^2p(\bm{r}, t)}{\partial t^2} = s(\bm{r}, t)\;,
\end{equation}
where $\nabla^2$ is the Laplace operator, $p(\bm{r}, t)$ denotes the pressure wavefield at spatial location $\bm{r}$ and time $t$, $v(\bm{r})$ represents the velocity map of wave propagation, and $s(\bm{r}, t)$ is the source term. As shown in Fig.~\ref{fig:fwi}, the goal of forward modeling is to simulate seismic data $\tilde{p}$ from a given velocity map $v$. For simplicity, we formulate this process as:
\begin{equation}
    \tilde{p} = f({v}),
\end{equation}
where $f(\cdot)$ represents the highly nonlinear forward operator. As mentioned above, data-driven FWI methods directly learn the inverse mapping as: 
\begin{equation}
\hat{v} = g_\theta(p)=f^{-1}({p}),
\end{equation}
where $\hat{v}$ is the estimated velocity map and $g_\theta(\cdot)$ is the approximated inverse operator of $f(\cdot)$, which is usually implemented as neural networks parameterized by $\theta$. BigFWI is developed to leverage large-scale datasets to obtain a more precise and universal approximation of the inverse operator.

\subsection*{Network Architecture} 
\pjremark{We introduce three variants of BigFWI, including a ``\textbf{B}ase'' model, a ``\textbf{M}iddle'' one with additional layers, and a ``\textbf{L}arge'' one that is both deeper and wider. We denote them as BigFWI-B, BigFWI-M and BigFWI-L. The number of the parameters of each model is summarized in Supplementary Table~S1. All BigFWI models share an encoder-decoder architecture.} The encoder $\mathcal{E}$ first extracts the spatial-temporal features from the seismic input $p\in\mathbb{R}^{S\times T\times R}$ and compresses them into a latent vector $z=\mathcal{E}(x)\in \mathbb{R}^{L\times 1\times 1}$. Here, $S$ equals the number of sources used in seismic surveys or simulation, $T$ represents the number of samples recorded by each receiver, $R$ denotes the number of receivers, and $L$ is the length of the latent vector. The decoder $\mathcal{D}$ then transforms the latent vector $z$ into spatial domain and generates the estimation of the velocity map $\hat{v}=\mathcal{D}(z)\in \mathbb{R}^{1\times W\times H}$, where $W$ and $H$ denote the horizontal (i.e. length) and vertical (i.e. depth) dimensions of the velocity map. Both the encoder $\mathcal{E}$ and the decoder $\mathcal{D}$ are fully based on 2D convolutional and deconvolution layers, and the details are presented as follows. The visualized network architecture of BigFWI is provided in Supplementary Figure~S1.

% encoder layers, decoder layers, latent length, # of parameters

In the encoder $\mathcal{E}$, since $T=1000$ is much larger than $R=70$ in the seismic data $p$ of \textsc{OpenFWI}, we first reduce temporal dimension and extract temporal features by stacking seven convolutional layers with $n\times1$ kernels, where $n=7$ in the first layer, and $n=3$ in the following six layers. The stride along the temporal dimension is set to 2 for every other layer to reduce the temporal dimension until it is close to the spatial dimension. We then stack six layers with $3\times3$ kernels to extract spatial-temporal features at the same time. Stride 2 is now applied to both dimensions every other layer. In BigFWI-M, instead of stacking six layers, we stack nine layers where an additional $3\times3$ layer with stride 1 is added after every two layers so as to increase model capacity without changing the dimensions of the original feature maps. In BigFWI-L, we stack eight layers where two additional $3\times3$ layer with 1024 features maps are appended at the end. In BigFWI-B and BigFWI-M, we use a layer with an $8\times9$ kernel to flatten the feature maps of the last to a 512-length latent vector $z$. In BigFWI-L, a layer with $4\times5$ kernel is used, and the length of latent vector is 1024. 

% \subsubsection*{CNN Decoder}
The decoder $\mathcal{D}$ includes five deconvolution layers for upsampling, and each of them is followed by one convolutional layer with $3\times3$ kernels in BigFWI-B and two convolutional layers in BigFWI-M. The first deconvolution layer with kernel size 5 transforms the latent vector $z$ into a $512\times5\times5$ tensor. The rest of the deconvolution layers with kernel size 4 and stride 2 upsample the feature maps by a factor of 2, resulting in an $80\times80\times32$ tensor. We then apply center cropping followed by a $3\times3$ convolutional layer to output a single channel $70\times70$ velocity map. In BigFWI-L, the kernel size of the first deconvolution layer is 2, and there are six groups of deconvolution and convolutional layers.

All the convolutional and deconvolution layers are followed by batch normalization and LeakyReLU as the activation function, except for the last output layer, which uses Tanh to generate the velocity map between $[-1, 1]$.

%  Loss function
\subsection*{Loss Function}
The original InversionNet model was trained with pixel-wise $\ell_1$ loss or $\ell_2$ loss between the ground truth of velocity maps $v$ and the predictions $\hat{v}$.  In this paper, we trained the baseline InversionNet and BigFWI using a combination of two loss functions to leverage the advantages from both sides according to the previous study~\cite{zhang2020data}. The loss function can be written as:
\begin{equation}
    \mathcal{L}(v, \hat{v}) = \frac{1}{W\cdot H} \sum_{i=1}^{W} \sum_{j=1}^{H}|v_{ij} - \hat{v}_{ij}| + \frac{1}{W\cdot H} \sum_{i=1}^{W} \sum_{j=1}^{H} \sqrt{(v_{ij} - \hat{v}_{ij})^2},
\end{equation}
where $W$ and $H$ denote the number of grids in horizontal length and depth directions, and $v_{ij}$ and $\hat{v}_{ij}$ represent the ground truth velocity and the prediction at the grid $(i,j)$.

\section*{Results}
\pjremark{In this section, we first describe the \textsc{OpenFWI} dataset and then present the evaluation metrics and training details, followed by the experimental results.} 

\subsection*{OpenFWI Datasets}
We here briefly describe the \textsc{OpenFWI} datasets which are used in all the experiments.
% All the experiments in our study are conducted on the \textsc{OpenFWI} datasets. 
Unlike many existing synthetic datasets for FWI, \textsc{OpenFWI} is publicly available and offers a rich collection of large-scale multi-structural benchmark datasets. 
% Families 
The datasets in \textsc{OpenFWI} are divided into four groups: ``\textit{Vel Family}'', ``\textit{Fault Family}'', ``\textit{Style Family}'' and ``\textit{Kimberlina Family}''. We exclude the ``\textit{Kimberlina Family}'' in our experiments because the dimensions of both velocity maps and seismic data in ``\textit{Kimberlina Family}" are different from the other three families. This allows us to combine the data samples from different datasets to train BigFWI models. 
% Naming
In terms of the complexity of subsurface structures, each of the three families consists of an easy version (\textit{-A}) and a hard version (\textit{-B}). In addition, the datasets in ``\textit{Vel Family}'' and ``\textit{Fault Family}'' are further divided into a flat version (\textit{Flat-}) and a curved version (\textit{Curve-}) in accordance with the shape of rock layers. The 10 datasets employed in our experiments are: \textit{FlatVel-A/B}, \textit{CurveVel-A/B}, \textit{FlatFault-A/B}, \textit{CurveFault-A/B}, and \textit{Style-A/B}. We use dataset abbreviations such as FVA for FlatVel-A in the rest of the paper to simplify plots. 

% Size and dimensions
Each dataset in ``\textit{Vel Family}", ``\textit{Fault Family}", ``\textit{Style Family}" is split into 24K/6K, 48K/6K, and 60K/7K pairs of seismic data and velocity maps for training and testing, respectively. We follow this splitting through our experiments. Figure~\ref{fig:fwi} shows an example of a velocity map and seismic data pair. Each velocity map has dimensions of $70\times70$ (depth $\times$ length in grids) with a grid spacing of 10 meters in both directions. The dimensions of the seismic data are $5\times1000\times70$ (\# of sources $\times$ \# of timesteps $\times$ \# of receivers). Five sources are evenly distributed on the top surface, each of which is a Ricker wavelet with a central frequency of 15 Hz. The interval between timesteps is 1 millisecond, and the receivers are also placed with an interval of 10 meters. For more details about the forward model algorithm and simulation, please refer to the original paper of \textsc{OpenFWI}.

% Additional data
% In addition to the samples in \textsc{OpenFWI}, we follow the data generation pipelines described in the original paper\cite{openfwi} and generate additional velocity maps and seismic data for each dataset, as mentioned in the previous section. % section name TBD

% evaluation
\subsection*{Evaluation Metrics}
We follow the benchmarking guidelines in \textsc{OpenFWI} and compute three metrics between the ground truth and the prediction of velocity maps to evaluate the performance of a model: mean absolute error (MAE), root mean squared error (RMSE), and structural similarity (SSIM)~\cite{wang2004image}. Both MAE and RMSE are commonly used to measure pixel-wise errors, while SSIM aligns better with human vision and measures the perceptual similarity that is more related to structural information. When calculating MAE and RMSE, we keep the velocity maps in the normalized scale $[-1, 1]$. During the calculation of SSIM, we rescale the velocity maps to $[0, 1]$ as required by the algorithm. \pjremark{We additionally compute the average quadratic Wasserstein Distance~\cite{yang2018application, bonneel2015sliced} for both velocity maps and seismic data as side evaluation metrics. The details are provided in Supplementary Wasserstein Distance Section.}

\subsection*{Training Details}
% training
We use identical hyperparameters to train all the models in our experiments. Specifically, we employ AdamW optimizers with momentum parameters $\beta_1=0.9$, $\beta_2=0.999$, and a weight decay of $1\times10^{-4}$ to update the parameters of each model. The base learning rate is set to $8\times10^{-4}$, and the models are trained for 170 epochs. In the first five warm-up epochs, we linearly increase the learning rate from $1\times10^{-4}$, and we decay the learning rate by a factor of 10 at epoch 150 and epoch 160, respectively. The batch size is set to 256. All the models are implemented in PyTorch and trained on 4 NVIDIA Tesla V100 GPUs. We employ the natural logarithmic transformation to make the intensity of seismic data more balanced and normalize the data to range $[-1,1]$ before they are fed into the network. The velocity maps are also normalized to the same scale before we compute the loss.

\subsection*{Big data benefits FWI} \label{sec:msg1}
% Shihang
We first design an experiment to explore if the performance of the data-driven FWI models can be improved by enlarging the training set. Specifically, we train a BigFWI-B on a combination of the datasets in \textsc{OpenFWI} and compare performance with the baseline InversionNet models trained on each split dataset. For brevity, we refer to BigFWI-B as BigFWI in this section. Both the BigFWI and InversionNet models share the same training hyperparameters and network architecture. Instead of using all the training samples in \textsc{OpenFWI}, we randomly select 12K samples from each of the four datasets in Vel Family, 24K from each of the four in the Fault Family, and 30K from each of the two in the Style Family. The rest of the samples in the training sets are reserved for the experiment in the next section, where we further enlarge the datasets. The combined large-scale training set consists of 204K samples in total, and we name it OpenFWI-204K. The test sets are directly adopted from \textsc{OpenFWI}.

\begin{figure}[t]
\centering
\includegraphics[width=\textwidth]{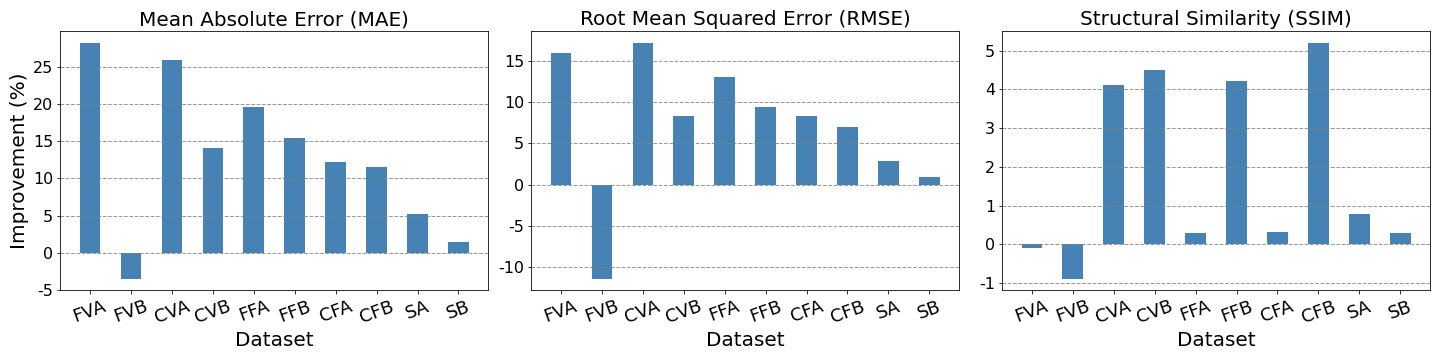}
\caption{\textbf{Performance improvement of BigFWI trained on OpenFWI-204K over InversionNet} in terms of MAE, RMSE and SSIM. BigFWI trained on a large-scale dataset (OpenFWI-204K) yields better performance on almost every dataset, compared to InversionNet~\cite{wu2019inversionnet}, which was trained on relatively small-scale datasets.}
\label{fig:204k_imp}
\end{figure}

\begin{figure}[t]
\centering
\includegraphics[width=\textwidth]{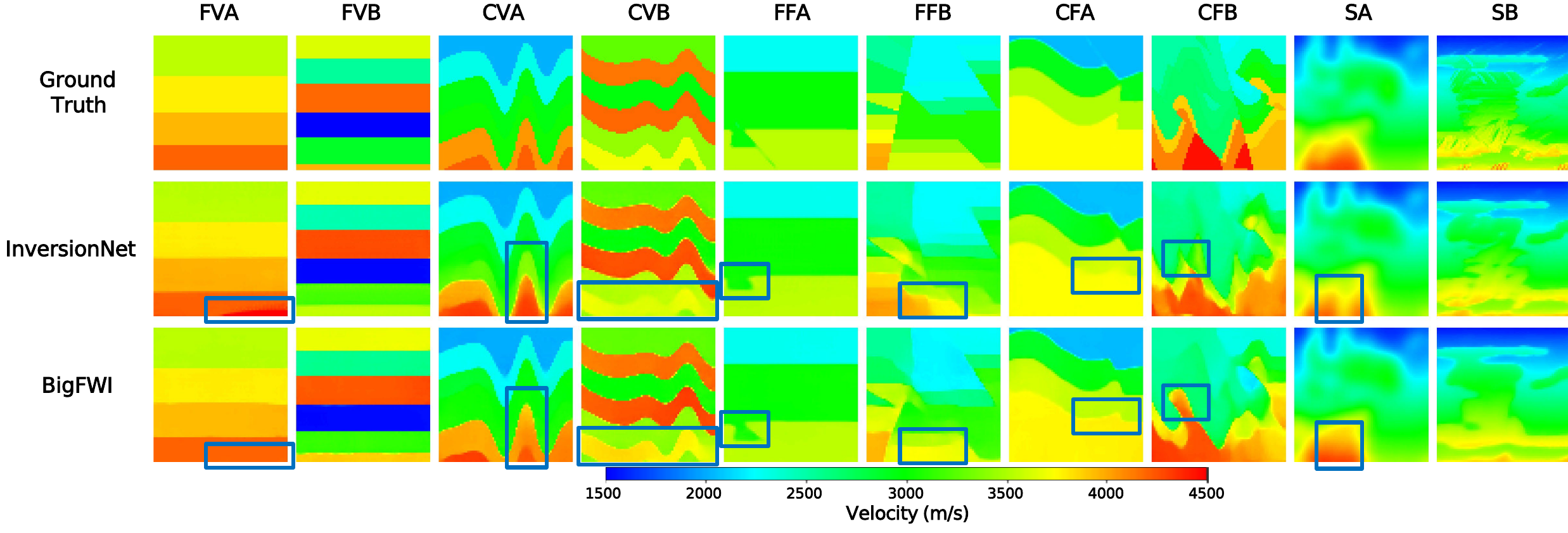}
\caption{\textbf{Comparison of ground truth (top) and predicted velocity maps generated by InversionNet (middle) and BigFWI (bottom)}. In general, BigFWI yields clearer layer boundaries, more accurate fault locations, and fewer artifacts compared to InversionNet as highlighted in squares.}
\label{fig:204k_vis}
\end{figure}

We plot the performance improvement of BigFWI compared to the InversionNet on each dataset in Figure~\ref{fig:204k_imp}. The quantitative results are provided in Supplementary Table~S2 \pjremark{and S3}. We observe that BigFWI shows a clear improvement for all the datasets except for datasets FVA and FVB, which are comprised of flat layers only. One potential reason for the model's degraded performance on FVA and FVB is that the network focuses more on curved layers which exist in most of the other datasets, and thus has a negative impact on the prediction of flat layers. We also observe that BigFWI exhibits significant improvement in MAE and RMSE for A datasets compared to B datasets across all families. However, the comparison of SSIM demonstrates the opposite trend, with the B datasets exhibiting better SSIM improvements compared to the A datasets in the same family. This variation in performance could be attributed to the greater complexity of the B datasets. The discrepancies in the baseline structures may not impact statistical misfits such as MAE and RMSE, but they may influence the SSIM. The simpler A datasets tend to benefit slightly more from the larger data volume than the more intricate B datasets.

Figure~\ref{fig:204k_vis} shows a comparison of velocities maps between ground truth, InversionNet, and BigFWI. We observe that InversionNet predicts the velocity maps with various errors, such as extra bottom layer anomalies (FVA), inaccurate layer values (CVA, CVB), and inaccurate structures (FVB, FFB, CFA, CFB). BigFWI models generally yield better performance in predicting the structure and values of the velocity maps than InversionNet. We see that the improved results of BigFWI are due to the knowledge learned from the 
large-scale training dataset that consists of a variety of velocity map distributions. Here, we define the velocity map distributions as the different geological subsurface structures in OpenFWI: i.e., flat layers vs. curved layers, faults vs. non-fault, and smooth vs. sharp. 
However, in addition to benefits, the variety of the velocity map distributions may also bring some negative effects such as inaccurate layer boundaries. For instance, we observe non-flat interfaces in the predictions of FVA/FVB, which are obviously affected by other velocity map distributions.

\pjremark{Additionally, we conduct an experiment by simply enlarging each split dataset, and this also leads to performance improvement. Results are provided in Supplementary Table S4 and S5, followed by a discussion in Supplementary Single Enlarged Dataset Section.}
%
% However, it is also because of the variety that BigFWI models may have structure layer boundary inaccuracy, such as the non-flat interfaces in FVA/FVB, which is transferred from other domains. This is consistent with the previous observations in Figure~\ref{fig:204k_imp}. In addition, since FVB contains more layers than FVA, the model performance on FVB degrades much more than it on FVA. 

% Why improvment of A larger than B in terms of MAE & RMSE?
% Why improvment of B larger than A in terms of SSIM?

% Qualitative analysis
% Peng's observation for reference
% FVA:
%  - InversionNet: bottom layer anomaly
%  - BigFWI: accurate bottom layer but layer boundary affected by other datasets
% FVB:
%  - InversionNet: bottom layer inaccurate
%  - BigFWI: same as FVA
% CVA  InversionNet: bottom layer blurry, inaccurate additional layer
% CVB  InversionNet: layer boundary blurry
% FFA  InversionNet: inaccurate fault in shallow layer
% FFB  InversionNet: inaccurate fault
% CFA  InversionNet: layer boundary blurry
% CFB  challenge for both, but BigFWI predicts more accurate layer boundaries
% SA   similar results
% SB   InversionNet: more high frequency component, visually better than BigFWI

\subsection*{Big data in FWI requires larger models}

\begin{figure}[t]
\centering
\includegraphics[width=\textwidth]{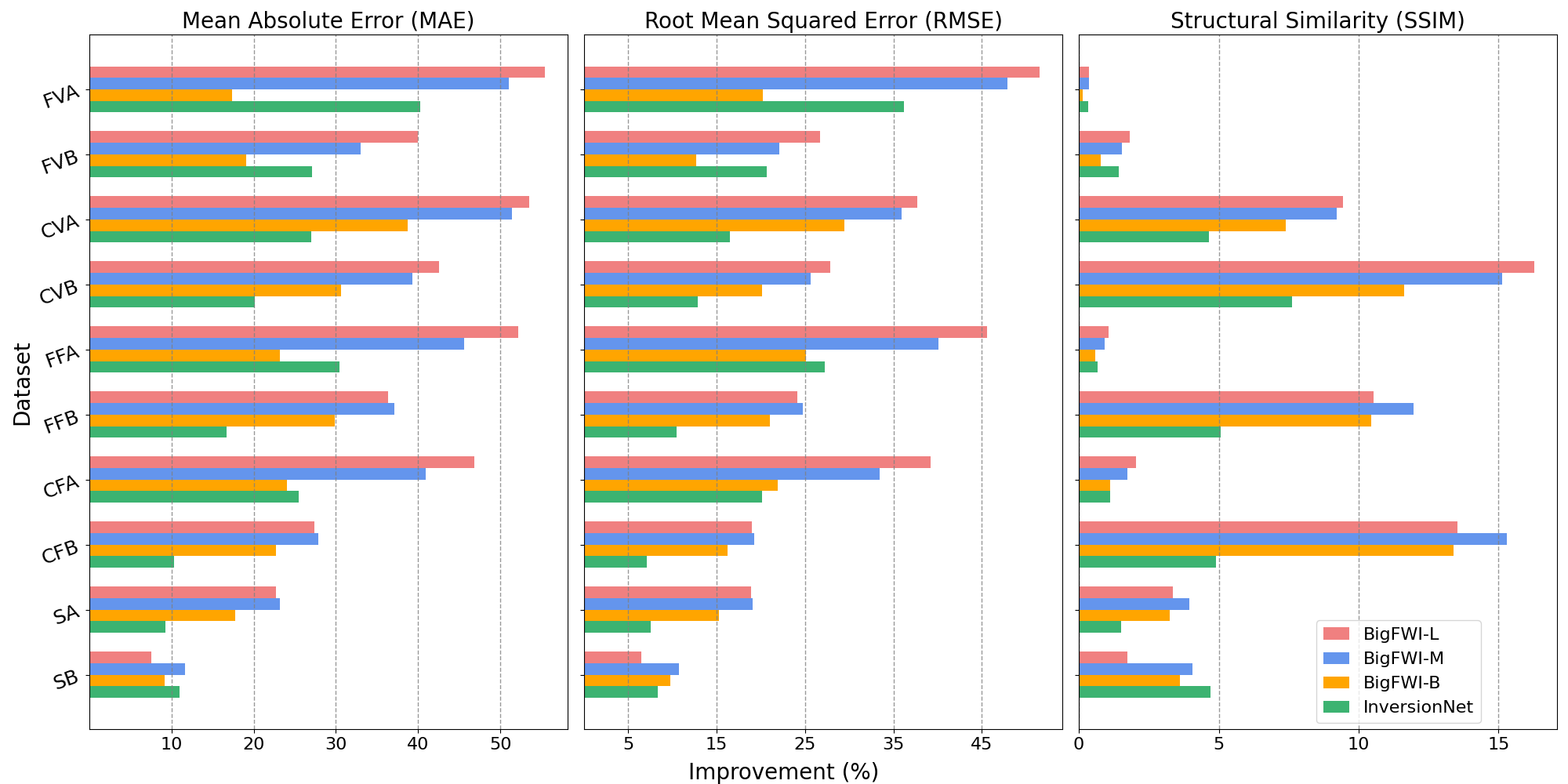}
\caption{\textbf{Comparison of the performance improvement of different methods trained on the further enlarged dataset (i.e. OpenFWI-408K)} in terms of MAE, RMSE, and SSIM. Note that the improvement percentages are computed based on the InversionNet trained on OpenFWI-204K. For most of the datasets, the BigFWI-L, which has the largest model size, yields the best performance.}
\label{fig:408k_imp}
\end{figure}

% Hanchen
To explore the relationship between the size of the training set and the size of the data-driven FWI models, we conduct an experiment that is similar to the previous one but employs the full training set provided by \textsc{OpenFWI}. Hence, the training set now contains 408K samples, and we name it OpenFWI-408K for brevity. We keep the baselines which are InversionNet trained on OpenFWI-204K, and we additionally train the InversionNet on each split datasets of OpenFWI-408K for comparison. We also train BigFWI-B, BigFWI-M and BigFWI-L on OpenFWI-408K.

% To investigate the necessity of employing larger data-driven FWI models for seismic imaging problems when the size of the training set is further enlarged. To achieve this, an experiment was designed that is similar to a previous one but employs all the samples in each officially split training set. Specifically, the training set now contains 408K samples, and we name it OpenFWI-408K for brevity. Additionally, the network architecture of the BigFWI model was expanded, making it deeper by adding additional convolutional layers and named BigFWI-L. The details of the expanded network architecture are discussed later in Section Network Architecture.

% \newline
In Figure~\ref{fig:408k_imp}, we show the statistical performance improvement of BigFWI over InversionNet (trained on the split components of OpenFWI-204K). More detailed quantitative results are provided in Supplementary Table~S\pjremark{6 and S7}. Overall, all the models trained on OpenFWI-408K yield better performance compared to InversionNet trained on OpenFWI-204K, and the BigFWI models (coral, blue and orange bars) outperform InversionNet (green) for almost every dataset, which again verifies that larger training set brings better performance. We further observe that among three BigFWI variants, larger models yield better performance in general. In particular, BigFWI-L (coral) and BigFWI-M (blue) outperform BigFWI-B (orange) by a large amount in all three metrics for relatively simple datasets such as FVA, FVB, FFA and CFA. For relatively complicated datasets such as CFB and SA, the gap is narrower. For dataset SB, BigFWI-B even outperforms BigFWI-L. This infers that larger models are preferred for most big data scenarios, but additional efforts such as more advanced network architectures are still required for some complicated cases.
% It is also likely that exposing the neural network to more complex training datasets would also improve generalization.

% We observe that the BigFWI-L model trained on OpenFWI-408K (blue bars) has the largest improvement in all three statistic matrices across all the datasets compared to the BigFWI (orange) and InversionNet (green) models trained with the same data volume.

% \newline

\begin{figure}[t]
\centering
\includegraphics[width=\textwidth]{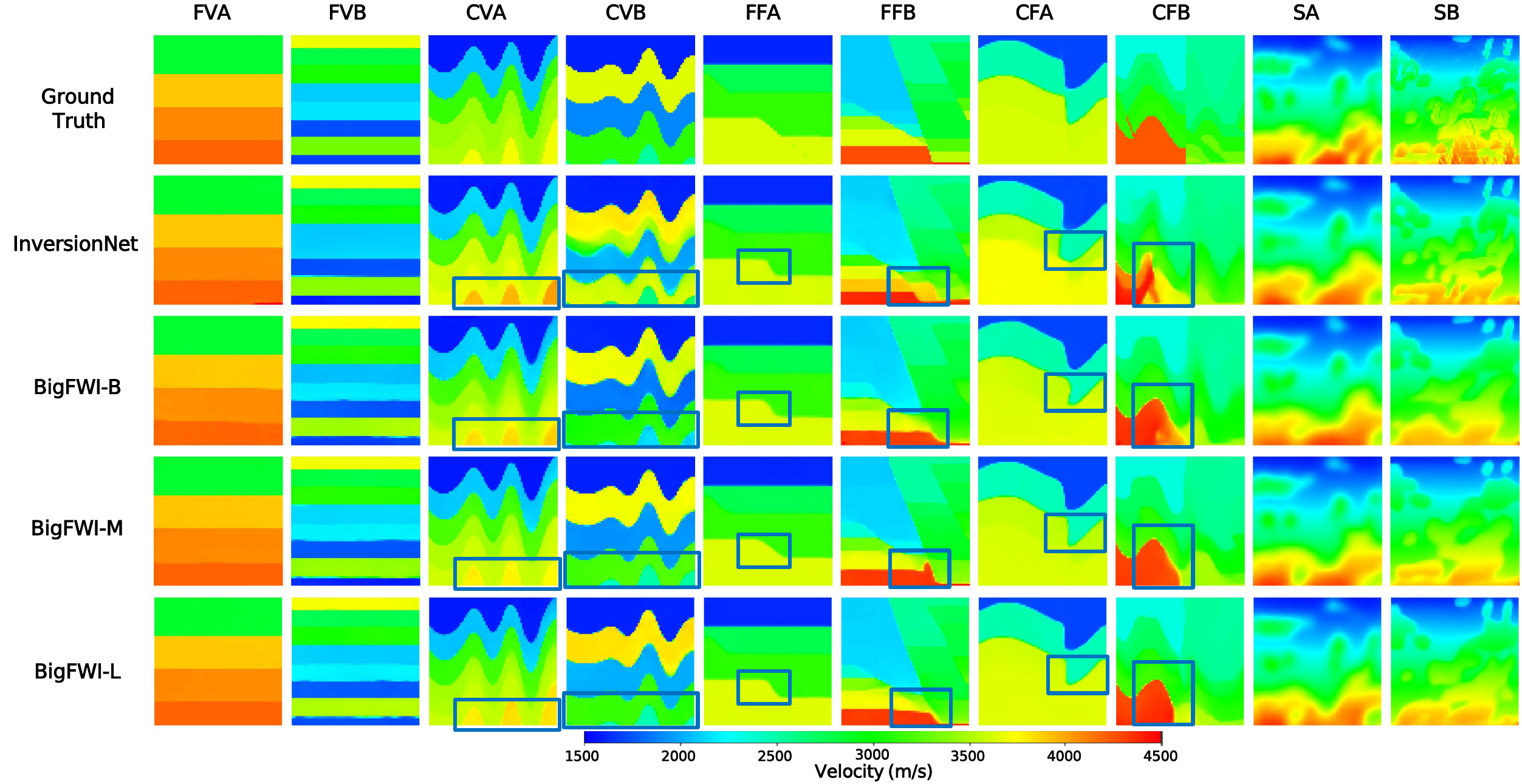}
\caption{\textbf{Comparison of ground truth (first row) and predicted velocity maps generated by InversionNet (second row) and BigFWIs (from third row to fifth row)}. BigFWI-L yields the most accurate results, especially in deep regions, compared to other BigFWI models and the baselines as highlighted in squares.}
\label{fig:408k_vis}
\end{figure}

% \newline
Figure~\ref{fig:408k_vis} shows the ground truth and predictions of velocity maps InversionNet, BigFWI-B, BigFWI-M, and BigFWI-L. Though the performance of InversionNet has improved statistically when trained on larger datasets, errors in prediction such as extra bottom layer anomalies (FVA), inaccurate layer values (CVA, CVB), and inaccurate structures (FVB, FFB, CFA, CFB) still exist. In contrast, BigFWI generally offers enhanced accuracy in layer location and velocity values. Comparing the performance of the BigFWI models, BigFWI-L and BigFWI-M outperforms BigFWI-B in many aspects. For instance, the flat interfaces in FVA and FVB are more flat and sharp in the results of BigFWI-L and BigFWI-M than the ones of BigFWI-B. BigFWI-M also predicts more accurate fault slopes in FFA and FFB. A similar observation can be obtained from the Style Family results, in which BigFWI-L and BigFWI-M predict more accurate kinematic information than InversionNet and BigFWI-B. Though InversionNet predicts more high-frequency components, the scatters are inaccurate in shape, which introduces even larger data misfit. 

% \newline
% When processing large volumes of seismic data and imaging the corresponding subsurface structures, large-scale networks are more effective than small-scale networks. The reason behind this is that the processing, analysis, and storage of seismic data can be quite challenging due to its volume and complexity. In this regard, large-scale neural networks have a distinct advantage over their smaller counterparts as they are better equipped to manage these challenges. Therefore, it is important to scale up the size of the neural network used for seismic imaging in proportion to the volume of seismic data being analyzed to obtain accurate results. As seismic data continues to increase in volume, small-scale neural networks may not be able to process, analyze, and image the data effectively. On the other hand, large-scale neural networks can handle the demands of processing and analyzing massive volumes of seismic data, enabling them to extract more detailed and accurate insights. In summary, to achieve optimal results in seismic imaging, it is essential to enlarge the scale of the neural network used in proportion to the volume of seismic data being processed.

% Why improvement of A larger than B in terms of MAE & RMSE?
% Why improvement of B larger than A in terms of SSIM?

% Qualitative analysis
% Peng's observation for reference
% FVA:
%  - InversionNet: bottom layer anomaly (small)
%  - BigFWI: no anomaly accurate bottom layer but layer boundary affected by other datasets
%  - BigFWI-L: more flat
% FVB:
%  - InversionNet: one layer missing
%  - BigFWI & BigFWI-L: not flat
% CVA  InversionNet: bottom layer inaccurate
% CVB  InversionNet: bottom layer inaccurate
% FFA  BigFWI-L: more accurate fault slope 
% FFB  InversionNet: extra layer on the left
% CFA  InversionNet: extra layer & inaccurate fault
% CFB  InversionNet: high velocity area at the bottom not complete
% SA   similar results
% SB   similar results, but InversionNet yields more high frequency components, visually better than BigFWI; BigFWI improved but still not enough

\subsection*{Big data leads to better generalization}
% Yinan
\textbf{Leave-one-out Generalization Test:} To verify whether large-scale training data also leads to better generalization, we design the experiment where the BigFWI models are trained under leave-one-out settings. Specifically, given a target dataset for testing (e.g., FVA), we train the BigFWI model on the combination of the training samples from all the other datasets in \textsc{OpenFWI} (e.g., FVB, CVA/B, FFA/B, CFA/B, and SA/B). We then compare the performance of this BigFWI model on the test samples of the target dataset (e.g., FVA) with InversionNet, which are trained on split datasets other than the target one. 

\begin{figure}[t]
\centering
\includegraphics[width=\textwidth]{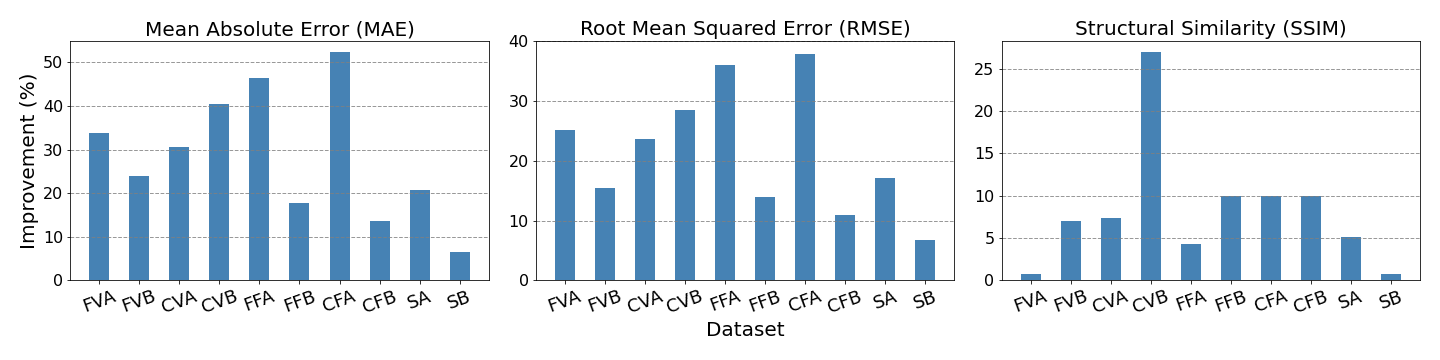}
\caption{\textbf{Generalization improvement of BigFWI models trained using leave-one-out settings} in terms of MAE, RMSE and SSIM. For each target dataset, our BigFWI yields better generalization performance than all the InversionNet trained on the split datasets other than the target one. }
\label{fig:gen}
\end{figure}

In Figure~\ref{fig:gen}, we present the statistical performance improvement in the percentage of the best generalization performance of InversionNet models. BigFWI shows superior performance across all the datasets, especially in terms of MAE and RMSE. This yields that big data leads to better generalization. Notably, utilizing datasets A as the target set results in greater improvements in terms of MAE and RMSE, while datasets B show greater improvements in terms of SSIM. The detailed quantitative results are provided in Supplementary Table~S\pjremark{8 and S9}.

Figure~\ref{fig:gen_vis} compares the generalization results of different methods to the ground truth. We observe that InversionNet produces inaccurate layer structures for out-of-distribution (OOD) data. In FFA, FFB, CVA, FFA, and SA, InversionNet's generalization outputs have errors of blurred borders, wrong layer positions, and inaccurate velocity values, especially in deeper parts. Moreover, the results clearly have incorrect patterns from other datasets in more complex datasets (i.e., CVB, FFB, CFA and CFB). Meanwhile, these explain why we could find higher SSIM improvement in these four datasets in Fig.~\ref{fig:gen}. Conversely, our BigFWI benefits from its large-scale cross-domain training set and can effectively capture more essential features of different datasets. Thus, BigFWI has more accurate predictions on OOD data than InversionNet.

\textbf{Generalization Test on Marmousi \& Overthrust:}
We further conduct generalization experiments on two more challenging standard test synthetic datasets Marmousi~\cite{brougois1990marmousi,martin2006marmousi2} and Overthrust~\cite{aminzadeh1996three, aminzadeh19963}. Both velocity maps contain more practical subsurface structures and have been widely adopted for the evaluation of full waveform inversion methods~\cite{feng2021multiscale, Jin-2021-Unsupervised, deng2022openfwi, feng2022intriguing}. Furthermore, the Marmousi velocity map was used as the style image to generate the Style Family in OpenFWI, which was specifically created for the simulation of real-world velocity maps. 

In this experiment, we resize the original Marmousi and Overthrust velocity maps to match our sizes and generate the seismic data using the same configuration as in OpenFWI. Since the dataset SA is a smoothed version of SB in OpenFWI, we are also interested in the generalization performance of BigFWI on the smoothed versions of Marmousi and Overthrust. To this end, we follow the previous work~\cite{feng2021multiscale} and apply Gaussian filters with a standard deviation 2 to the velocity maps to obtain the smoothed ones. For comparison, we compare the BigFWIs trained on OpenFWI-408K with the InversionNet models trained on SA and SB separately. 

\begin{figure}[t]
\centering
\includegraphics[width=\textwidth]{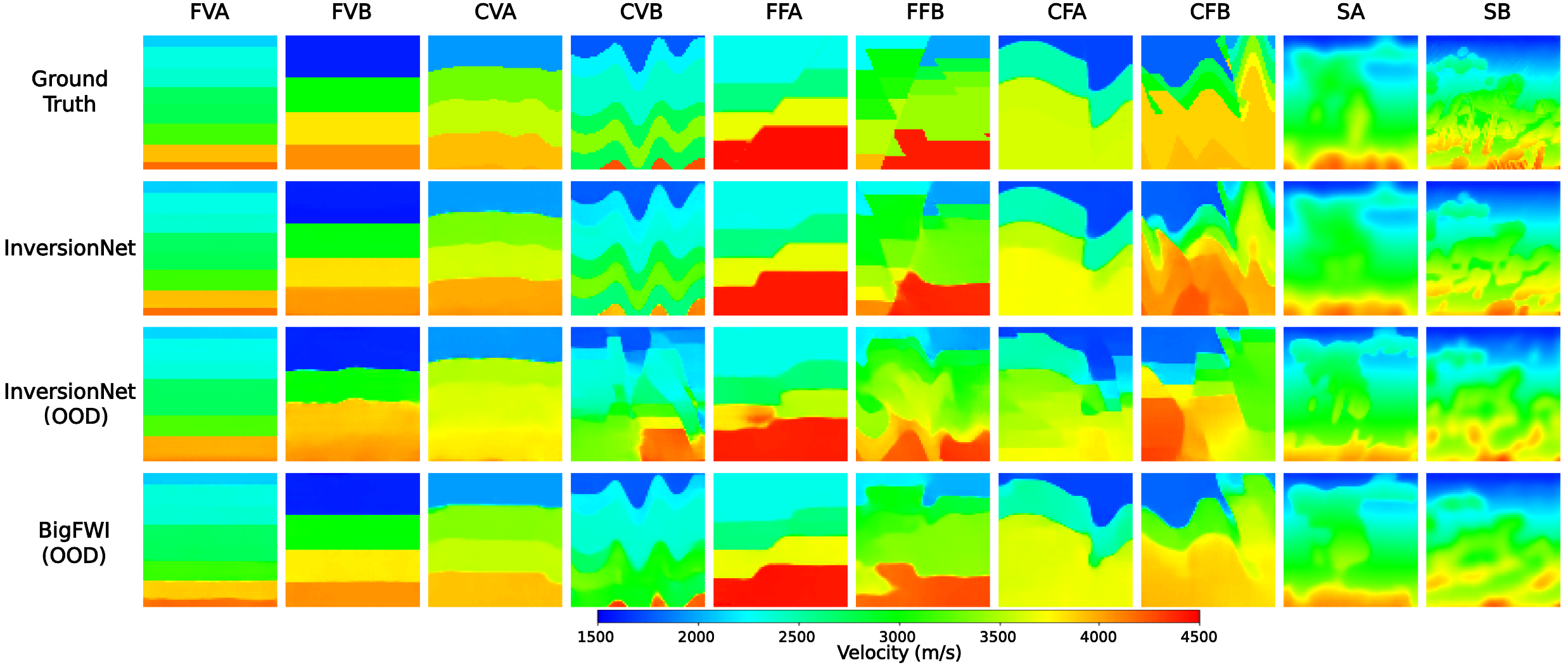}
% Comparison of ground truth (first row) and predicted velocity maps generated by baselines (second row), BigFWI (third row) and BigFWI-L (fourth row)}. From the first to last row: the baselines trained on the target datasets, the baselines trained on the datasets other than the target one, and the BigFWIs trained using leave-one-out settings. 
\caption{\textbf{Comparison of ground truth (first row) and generalization results of different methods.} From the first to last row: the InversionNet trained on the target datasets, the InversionNet trained on the datasets other than the target one, and the BigFWIs trained using leave-one-out settings. Our BigFWIs yield relatively reasonable velocity maps that are closer to the ones generated by the InversionNet trained on the target datasets.}
\label{fig:gen_vis}
\end{figure}

\begin{figure}[t]
\centering
\includegraphics[width=0.8\textwidth]{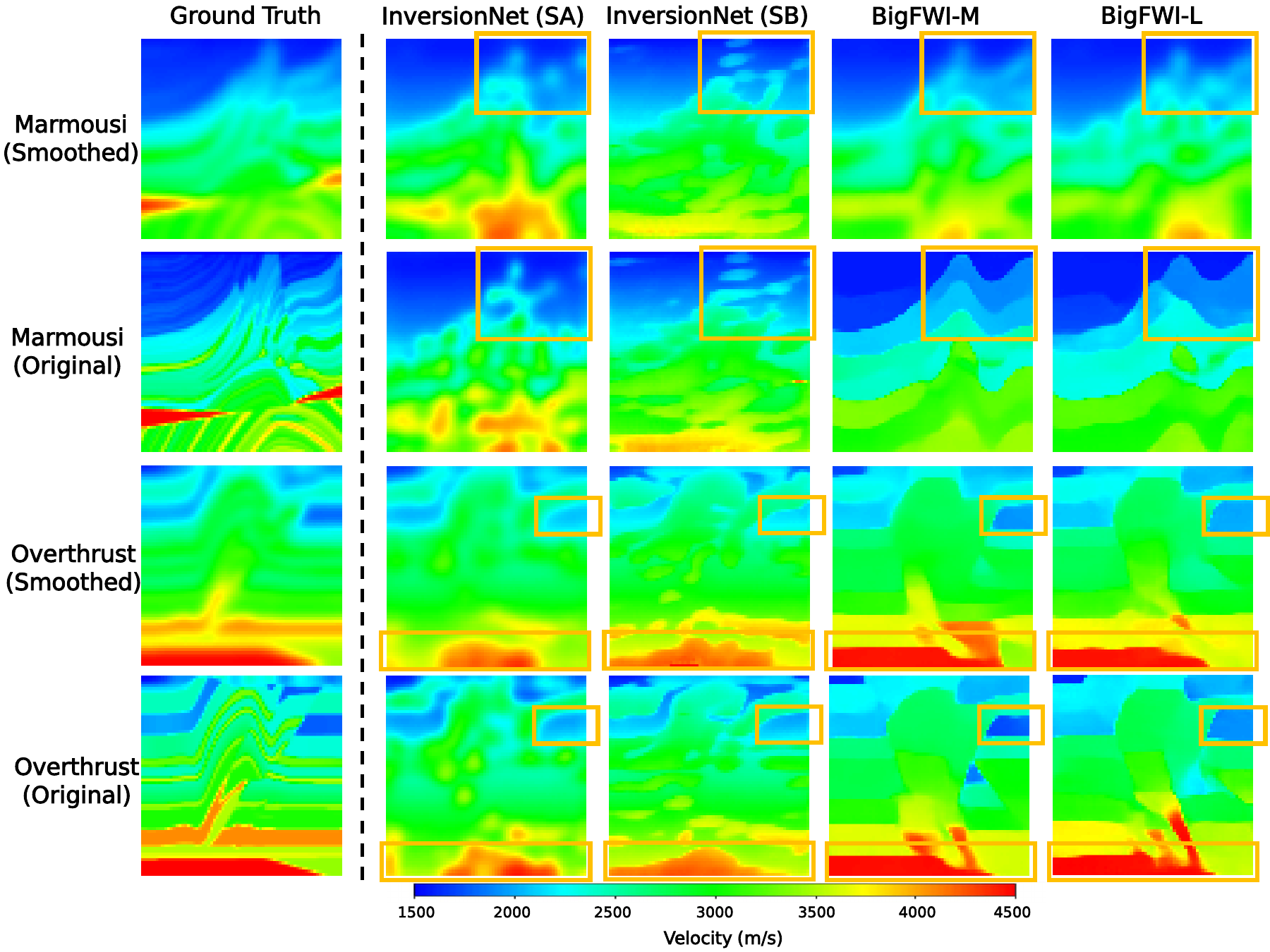}
% Comparison of ground truth (first row) and predicted velocity maps generated by baselines (second row), BigFWI (third row) and BigFWI-L (fourth row)}. From the first to last row: the baselines trained on the target datasets, the baselines trained on the datasets other than the target one, and the BigFWIs trained using leave-one-out settings. 
\caption{\textbf{Comparison of ground truth (first column) and generalization results of different methods on Marmousi and Overthrust.} For all velocity maps, Our BigFWIs yield more accurate results, especially in the shallow region of each velocity map as highlighted in squares.}
\label{fig:gen_marm_over}
\end{figure}

The generalization ability of our models to Marmousi and Overthrust are depicted in Figure~\ref{fig:gen_marm_over}. \pjremark{We also provide the results of Reverse Time Migration (RTM) and the differences of RTM compared to the ground truth in Supplement Figure S2 and S3, respectively.} Generally, BigFWI yield more accurate inversion results compared to InversionNet. For the smoothed version of Marmousi, the results of BigFWI match the ground truth better in the shallow region. The BigFWI-M even generates some layered structures in the top-right corner. In the deep region, the results of the InversionNet models contain either too many false high-velocity predictions or a horizontal layer with relatively low velocity. In contrast, though the velocity in the results of BigFWI is lower than the ground truth, they capture the locations of high-velocity regions. For the original version of Marmousi, it is obvious that the performance of BigFWI is better than InversionNet. We observe the layered structures given by BigFWI, and we think this is learned from CVA and CVB. 

For the Overthrust velocity maps, BigFWI consistently generates flat layers with geological faults in the deep region, which are more visually plausible than the results of InversionNet. We see that the behavior of BigFWI is greatly affected by FVB, which demonstrates the advantages of training models on large-scale multi-structural datasets. However, we also observe that BigFWI tends to follow one specific learned pattern per prediction; for example, in the predictions of the smoothed Overthrust, BigFWI still generates structures with sharp boundaries that exist in FVB. This indicates that there may be still much space for the improvement of BigFWI in terms of both the model architecture and the training data. 

\pjremark{The quantitative results are provided in Supplementary Table~S10, S11 and S12, which generally align with our observations in the visualization results. From the quantitative results, we further notice that the performance improvement of BigFWI compared to InversionNet on the original velocity maps is smaller than the one on the smoothed version. This is consistent with our previous observations where the improvement of BigFWI models on SB is always smaller than the one of SA. It points out a future direction where instead of simply combining all the datasets, we may bias towards SB dataset during training by generating more samples or training more steps on SB so that the model can yield better performance on the realistic cases with high-wavenumber components. In addition, we observe that InversionNet trained on SA achieves smaller RTM differences for the smoothed version of Overthrust. The discrepancy in this case may be attributed to a velocity misfit, causing RTM image interfaces to be half-cycle shifted in depth, resulting in larger RMS and L2-norm values. However, the performance of the four models on the Overthrust-smooth is relatively comparable. Moreover, it is worth noting that although BigFWI achieves better results compared to InversionNet for both Marmousi and Overthrust, the performance is still insufficient for real-world applications, which indicates much space for improvement.}

% The geometry of the Marmousi velocity model is based on a profile of the North Quenguela through the Cuanza basin~\cite{versteeg1993sensitivity}. It was designed to simulate a continental drift geological setting. Several typical geological structures were included, such as interfaces, faults, and strong velocity variations in both the lateral and the vertical direction.

\section*{Discussion}
% Peng
% We presented an empirical study to determine the extent to which big data can benefit the deep learning models in FWI from three perspectives: model performance, the relationship between model size and data size, and model generalization. To accomplish this, we utilized the large-scale, publicly available datasets \textsc{OpenFWI} and designed the experiments to compare the performance of baseline InversionNet trained on relatively small-scale individual datasets with that of BigFWIs, which are trained on combined, large-scale datasets. Through both quantitative and qualitative analysis, our study has demonstrated that big data can significantly enhance the performance of deep learning models in FWI on both in-distribution and out-of-distribution data. Moreover, we have shown that model architectures need to be scaled with data size to achieve further improvement. We trust that our findings can provide valuable guidance for the future development of deep-learning-based FWI methods.  

This study is a preliminary investigation into the influence of big data on deep learning FWI methods, and there still exist some limitations and promising future directions. First, our study is entirely based on \textsc{OpenFWI}, which brings us not only convenience but also several inherent limitations. Although the Style Family in \textsc{OpenFWI} has made an effort to simulate the real-world velocity maps, there is still a gap between the synthetic data and field data. Our experiments are thus limited to simulations. It is an ongoing challenge for the whole FWI community to bridge this gap by either providing more public field data or improving the fidelity of the simulation. Second, in the present study, we only made minimal modifications to the network architecture of BigFWI. As a potential direction of future work, we may develop different network architectures to further improve performance. For instance, we observed during the qualitative analysis that the cross-domain training could lead to interference between datasets and inaccurate layer boundaries. Such issues could potentially be addressed by implementing an adaptive network architecture. \pjremark{Another challenge we plan to take into consideration when developing the network architecture is how models can be generalized to various survey settings. Examples can be the size of the target velocity map, the type of source and the placement of source-receiver arrays. An existing method we can employ is Fourier-DeepONet~\cite{zhu2023fourier}, which considers the generalization of frequencies and locations of sources. Another potential solution is to adapt Transformer~\cite{vaswani2017attention} or Vision Transformer~\cite{dosovitskiy2020image} with embeddings that encode the additional information. If this challenge is well-addressed, we will be able to include more diverse data during training, and the model will be more easily generalized to practical field applications.} Third, throughout the experiments, we observe that the evaluation of inversion results is very complicated, and sometimes the differences in visualization results cannot be reflected in the current quantitative metrics. Hence, new evaluation metrics should also be developed in future to better reflect inversion quality.  % fixed input size may be solved by transformer

This study offers valuable insights into the inverse problem, which can contribute to the advancement of this concept in other domains, including medical imaging, climate modeling, and astronomy. The knowledge gained from this investigation can be leveraged to support the application of AI in scientific research and enhance its capabilities in these fields.

% The Discussion should be succinct and must not contain subheadings.
% Limitation & Future work
% synthetic datasets, challenge to obtain real data
% how to deal with datasets with different settings (# of sources, grid size, etc.)
% how to modify the network to condition it on the dataset (e.g. embedding)

\section*{Conclusion}
\pjremark{We presented an empirical study to determine the extent to which big data can benefit the deep learning models in FWI from three perspectives: model performance, the relationship between model size and data size, and model generalization. To accomplish this, we utilized the large-scale, publicly available datasets \textsc{OpenFWI} and designed the experiments to compare the performance of baseline InversionNet trained on relatively small-scale individual datasets with that of BigFWIs, which are trained on combined, large-scale datasets. Through both quantitative and qualitative analysis, our study has demonstrated that big data can significantly enhance the performance of deep learning models in FWI on both in-distribution and out-of-distribution data. Moreover, we have shown that model architectures need to be scaled with data size to achieve further improvement. We trust that our findings can provide valuable guidance for the future development of deep-learning-based FWI methods.}

% \section*{Data Availability}

% \section*{Code Availability}

\section*{Data and Codes Availability}
OpenFWI data set can be downloaded from the website~(\url{https://openfwi-lanl.github.io/}). InversionNet codes are released and can be downloaded from the Website~(\url{https://github.com/lanl/OpenFWI/}).

\bibliography{ms}

\section*{Acknowledgements}

This work was funded by the U.S. Department of Energy~(DOE) Office of Fossil Energy’s Carbon Storage Research Program via the Science-Informed Machine Learning to Accelerate Real Time Decision Making for Carbon Storage~(SMART-CS) Initiative.

\section*{Author contributions statement}

Y.C. and P.J. proposed the idea and initialized the project. P.J. conducted the experiments and plotted the figures. S.F. and H.W. contributed geophysical expertise. P.J., Y.F., S.F., and H.W. analyzed the results and wrote the manuscript. Y.C., B.C., and Y.L. provided comments on the manuscript. Y. L. and Y.C. supervised the project.

\section*{Competing interests}

The authors declare no competing interests.

% To include, in this order: \textbf{Accession codes} (where applicable); \textbf{Competing interests} (mandatory statement). 

% The corresponding author is responsible for submitting a \href{http://www.nature.com/srep/policies/index.html#competing}{competing interests statement} on behalf of all authors of the paper. This statement must be included in the submitted article file.

\end{document}

\maketitle

\section*{Supplementary}
\renewcommand{\thetable}{S\arabic{table}}
\renewcommand{\thefigure}{S\arabic{figure}}

\color{black}
\subsection*{Wasserstein Distance}
We follow the previous literature~\cite{yang2018application} to calculate the squared 2-Wasserstein Distance. The definition of the distance between two distributions $f:X\rightarrow \mathbb{R}^+$ and $g:Y\rightarrow \mathbb{R}^+$ can be formulated as: 
\begin{equation}
    W_2^2(f, g)=\inf_{T\in \mathcal{M}}\int\limits_{X} |x-T(x)|^2 f(x)dx,
\end{equation}
where $f$ and $g$ denotes the probability density functions of the distributions, $T:X\rightarrow Y$ is a transport plan, and $\mathcal{M}$ is the set of all possible transport plans.

For seismic data, we compute the average trace-by-trace distance. Each trace is considered as a density function. Additionally, we extract the envelope of data by applying the Hilbert transform to generate non-negative density functions. We also normalize the data of both ground truth and predictions so that the functions have equal mass, as required by Wasserstein Distance. We use $\tilde{f}$ and $\tilde{g}$ to denote the modified traces of the ground truth and the predictions. The average trace-by-trace distance is given as:
\begin{equation}
    \mathrm{WD}_{Seis}= \frac{1}{S\cdot R}  \sum_{s=1}^{S} \sum_{r=1}^{R} W_2^2(\tilde{f}(r,t;s), \tilde{g}(r,t;s)),
\end{equation}
where $S$ is the number of sources, and $R$ is the number of receivers.

Similarly, for velocity maps, we consider a map as a joint probability density function and estimate the distances by computing the Sliced Wasserstein Distances with 50 projections~\cite{bonneel2015sliced}.

\subsection*{Enlarged Single Dataset}
We further conduct an experiment by simply enlarging the split datasets in OpenFWI. Specifically, we follow the velocity map generation process in OpenFWI and generate additional pairs of velocity maps and seismic data for the \textit{Vel Family} and the \textit{Fault Family} so that each split dataset contains 204K training samples. The test sets remain unchanged. For brevity, we name the datasets as OpenFWI-204K-Extended, and the models trained on the enlarged datasets are still referred as BigFWI for consistency. For the \textit{Style Family}, there exist some factors that may cause distribution shift. For instance, the specific Marmousi velocity map used as the style image is unknown. Therefore, we choose to exclude the \textit{Style Family} in this experiment. The hyperparameter setting is the same as the one in the experiment of OpenFWI-204K. 

The quantitative results are listed in Table~\ref{tab:204k-se} and \ref{tab:204k-se-wd}. We observe that BigFWI outperforms InversionNet on all the datasets by a large extent, which still demonstrates the impact of data scaling. It is also worth noting that the performance improvement from enlarging single datasets is larger than the one of combining all the datasets. However, training on the single dataset may also limit the generalization of the model as the velocity maps with other subsurface structures are unseen during training.

% whether compare to the results OpenFWI-408k or not

\begin{figure}[t]
\centering
\includegraphics[width=\textwidth]{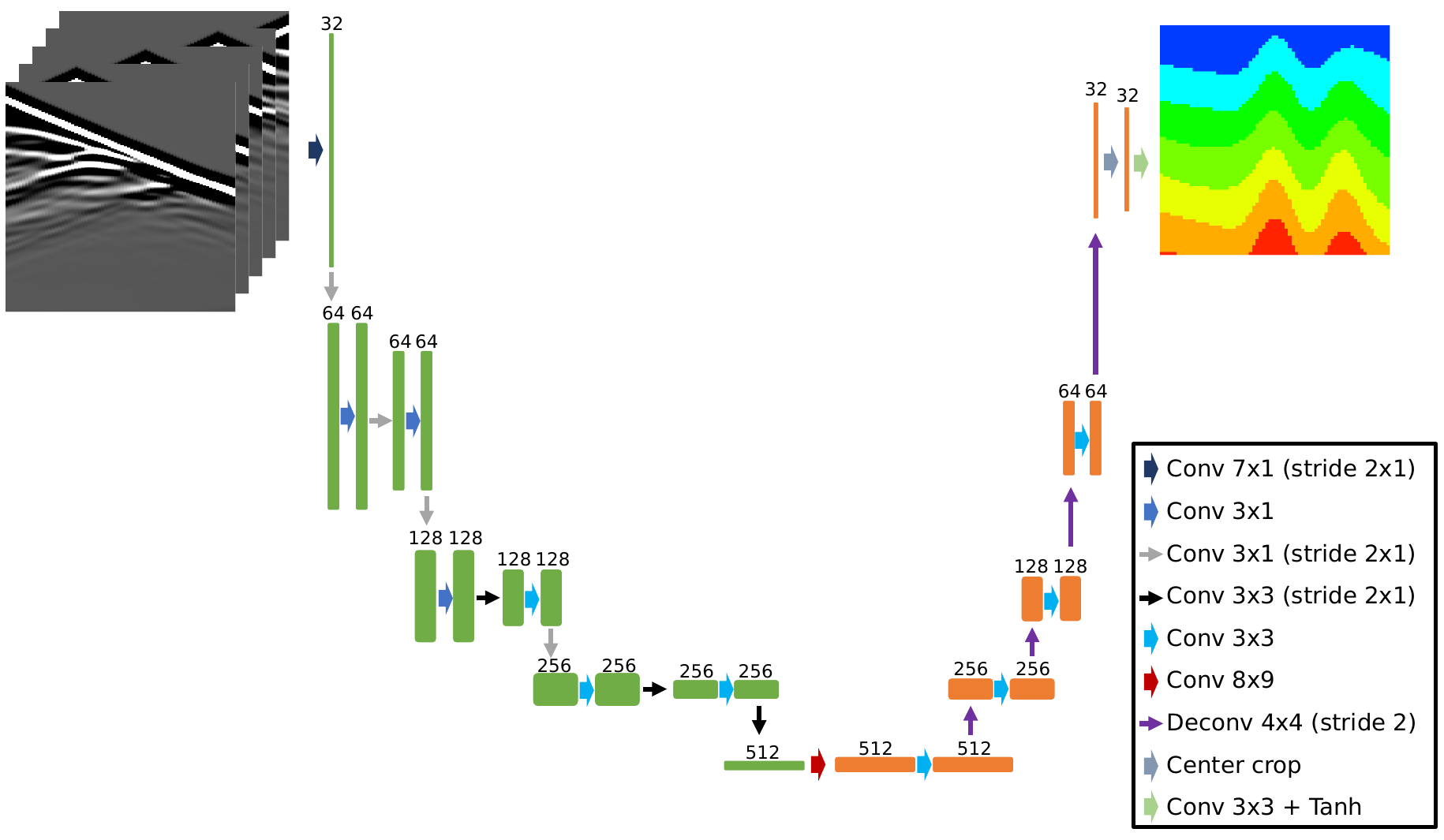}
\caption{Visualization of the network architecture of InversionNet and BigFWI. Batch normalization layers and activation layers are omitted for simplicity. }
\label{fig:newtork_arch}
\end{figure}

\begin{table}[t]
\centering
\setlength{\tabcolsep}{6pt}
\renewcommand{\arraystretch}{1.25}
\begin{tabular}{c c c c c }
\thickhline
% Model & \makecell{Encode\\Layers} & \makecell{Decode\\Layers} & \makecell{Latent\\Length} & Parameters 
Model & Encode Layers & Decode Layers & Latent Length & Parameters \\ \thickhline
BigFWI-B & 14 & 11 & 512 & 24M \\ \hline
BigFWI-M & 17 & 16 & 512 & 28M \\ \hline
BigFWI-L & 16 & 14 & 1024 & 87M \\ \thickhline
\end{tabular}
\caption{Configurations of BigFWI model variants.}
\label{tab:model}
\end{table}

\begin{table}[t]
\centering
\setlength{\tabcolsep}{10pt}
\renewcommand{\arraystretch}{1.5}
\begin{tabular}{c|c c c|c c c}
\thickhline
\multirow{2}{*}{Dataset} & \multicolumn{3}{c|}{InversionNet} & \multicolumn{3}{c}{BigFWI} \\ \cline{2-7} 
& MAE$\downarrow$ & RMSE$\downarrow$ & SSIM$\uparrow$  & MAE$\downarrow$ & RMSE$\downarrow$ & SSIM$\uparrow$ \\ \thickhline
FlatVel-A  & 0.0092 & 0.0163 & \textbf{0.9930} & \textbf{0.0066} & \textbf{0.0137} & 0.9922 \\ \hline
FlatVel-B  & \textbf{0.0288} & \textbf{0.0797} & \textbf{0.9582} & 0.0298 & 0.0888 & 0.9498 \\ \hline
CurveVel-A  & 0.0560 & 0.1131 & 0.8405 & \textbf{0.0415} & \textbf{0.0936} & \textbf{0.8751}  \\ \hline
CurveVel-B  & 0.1344 & 0.2697 & 0.6995 & \textbf{0.1155} & \textbf{0.2471} & \textbf{0.7309}  \\ \hline
FlatFault-A  & 0.0138 & 0.0382 & 0.9813 & \textbf{0.0111} & \textbf{0.0332} & \textbf{0.9841}  \\ \hline
FlatFault-B  & 0.1012 & 0.1672 & 0.7267 & \textbf{0.0855} & \textbf{0.1514} & \textbf{0.7573}  \\ \hline
CurveFault-A  & 0.0220 & 0.0607 & 0.9605 & \textbf{0.0193} & \textbf{0.0556} & \textbf{0.9636}  \\ \hline
CurveFault-B  & 0.1610 & 0.2419 & 0.5981 & \textbf{0.1425} & \textbf{0.2251} & \textbf{0.6292}  \\ \hline
Style-A  & 0.0625 & 0.1024 & 0.8839 & \textbf{0.0592} & \textbf{0.0994} & \textbf{0.8908}  \\ \hline
Style-B & 0.0609 & 0.0971 & 0.7303 & \textbf{0.0600} & \textbf{0.0962} & \textbf{0.7324}  \\ \thickhline
\end{tabular}
\caption{Quantitative comparison between the results of InversionNet and BigFWI on OpenFWI-204k in terms of MAE, RMSE and SSIM.}
\label{tab:204k}
\end{table}

\begin{table}[t]
\centering
\setlength{\tabcolsep}{10pt}
\renewcommand{\arraystretch}{1.5}
\begin{tabular}{c|c c|c c}
\thickhline
\multirow{2}{*}{Dataset} & \multicolumn{2}{c|}{InversionNet} & \multicolumn{2}{c}{BigFWI} \\ \cline{2-5} 
 & WD$_{Seis}\downarrow$ & WD$_{Vmap}\downarrow$  & WD$_{Seis}\downarrow$ & WD$_{Vmap}\downarrow$  \\ \thickhline
FlatVel-A  & 15.8109 & 0.0417 & \textbf{10.4228} & \textbf{0.0372} \\ \hline
FlatVel-B  & 79.0238 & 0.1339 & \textbf{76.6044} & \textbf{0.1328} \\ \hline
CurveVel-A & 478.6276 & 0.1898 & \textbf{291.5563} & \textbf{0.1192} \\ \hline
CurveVel-B & 997.6450 & 0.4120 & \textbf{819.0640} & \textbf{0.3354} \\ \hline
FlatFault-A & \textbf{150.4610} & 0.0481 & 170.1272 & \textbf{0.0436} \\ \hline
FlatFault-B & 722.4929 & 0.3139 & \textbf{544.0522} & \textbf{0.2260} \\ \hline
CurveFault-A & 264.9666 & 0.0692 & \textbf{260.3809} & \textbf{0.0577} \\ \hline
CurveFault-B & 1342.0856 & 0.4040 & \textbf{992.1356} & \textbf{0.3091} \\ \hline
Style-A & 156.1971 & 0.1967 & \textbf{145.9914} & \textbf{0.1691} \\ \hline
Style-B & \textbf{131.8703} & 0.0994 & 182.7130 & \textbf{0.0934} \\ \thickhline
\end{tabular}
\caption{\color{black}Quantitative comparison between the results of InversionNet and BigFWI on OpenFWI-204k in terms of Wasserstein Distance.}
\label{tab:204k-wd}
\end{table}

\begin{table}[t]
\centering
\setlength{\tabcolsep}{10pt}
\renewcommand{\arraystretch}{1.5}
\begin{tabular}{c|c c c|c c c}
\thickhline
\multirow{2}{*}{Dataset} & \multicolumn{3}{c|}{InversionNet} & \multicolumn{3}{c}{BigFWI} \\ \cline{2-7} 
& MAE$\downarrow$ & RMSE$\downarrow$ & SSIM$\uparrow$  & MAE$\downarrow$ & RMSE$\downarrow$ & SSIM$\uparrow$ \\ \thickhline
FlatVel-A  & 0.0092 & 0.0163 & 0.9930 & \textbf{0.0035} & \textbf{0.0056} & \textbf{0.9985} \\ \hline
FlatVel-B  & 0.0288 & 0.0797 & 0.9582 & \textbf{0.0116} & \textbf{0.0421} & \textbf{0.9867} \\ \hline
CurveVel-A  & 0.0560 & 0.1131 & 0.8405 & \textbf{0.0236} & \textbf{0.0703} & \textbf{0.9198}  \\ \hline
CurveVel-B  & 0.1344 & 0.2697 & 0.6995 & \textbf{0.0704} & \textbf{0.1953} & \textbf{0.8209}  \\ \hline
FlatFault-A  & 0.0138 & 0.0382 & 0.9813 & \textbf{0.0048} & \textbf{0.0196} & \textbf{0.9952}  \\ \hline
FlatFault-B  & 0.1012 & 0.1672 & 0.7267 & \textbf{0.0598} & \textbf{0.1288} & \textbf{0.8366}  \\ \hline
CurveFault-A  & 0.0220 & 0.0607 & 0.9605 & \textbf{0.0080} & \textbf{0.0324} & \textbf{0.9910}  \\ \hline
CurveFault-B  & 0.1610 & 0.2419 & 0.5981 & \textbf{0.1132} & \textbf{0.1964} & \textbf{0.6937}  \\ \hline
\end{tabular}
\caption{\color{black}Quantitative comparison between the results of InversionNet and BigFWI on OpenFWI-204k-Extended in terms of MAE, RMSE and SSIM.}
\label{tab:204k-se}
\end{table}

\begin{table}[t]
\centering
\setlength{\tabcolsep}{10pt}
\renewcommand{\arraystretch}{1.5}
\begin{tabular}{c|c c|c c}
\thickhline
\multirow{2}{*}{Dataset} & \multicolumn{2}{c|}{InversionNet} & \multicolumn{2}{c}{BigFWI} \\ \cline{2-5} 
 & WD$_{Seis}\downarrow$ & WD$_{Vmap}\downarrow$  & WD$_{Seis}\downarrow$ & WD$_{Vmap}\downarrow$  \\ \thickhline
FlatVel-A  & 15.8109 & 0.0417 & \textbf{2.6449} & \textbf{0.0355} \\ \hline
FlatVel-B  & 79.0238 & 0.1339 & \textbf{16.8437} & \textbf{0.0634} \\ \hline
CurveVel-A & 478.6276 & 0.1898 & \textbf{122.2878} & \textbf{0.0859} \\ \hline
CurveVel-B & 997.6450 & 0.4120 & \textbf{393.6323} & \textbf{0.2119} \\ \hline
FlatFault-A & 150.4610 & 0.0481 & \textbf{66.2190} & \textbf{0.0284} \\ \hline
FlatFault-B & 722.4929 & 0.3139 & \textbf{276.2333} & \textbf{0.1632} \\ \hline
CurveFault-A & 264.9666 & 0.0692 & \textbf{119.6634} & \textbf{0.0381} \\ \hline
CurveFault-B & 1342.0856 & 0.4040 & \textbf{696.8380} & \textbf{0.2314} \\ \hline
\end{tabular}
\caption{\color{black}Quantitative comparison between the results of InversionNet and BigFWI on OpenFWI-204k-Extended in terms of Wasserstein Distance.}
\label{tab:204k-se-wd}
\end{table}

\begin{table}[t]
\small
\centering
\setlength{\tabcolsep}{5pt}
\renewcommand{\arraystretch}{1.5}
\begin{tabular}{c|c c c|c c c|c c c|c c c}
\thickhline
\multirow{2}{*}{Dataset} & \multicolumn{3}{c|}{InversionNet} & \multicolumn{3}{c|}{BigFWI-B} & \multicolumn{3}{c|}{BigFWI-M} & \multicolumn{3}{c}{BigFWI-L} \\ \cline{2-13} 

& MAE$\downarrow$ & RMSE$\downarrow$ & SSIM$\uparrow$  & MAE$\downarrow$ & RMSE$\downarrow$ & SSIM$\uparrow$ & MAE$\downarrow$ & RMSE$\downarrow$ & SSIM$\uparrow$ & MAE$\downarrow$ & RMSE$\downarrow$ & SSIM$\uparrow$ \\ \thickhline

FlatVel-A & 0.0055 & 0.0104 & 0.9964  & 0.0076 & 0.0130 & 0.9943  & 0.0045 & 0.0085 & 0.9965 & \textbf{0.0041} & \textbf{0.0079} & \textbf{0.9965} \\ \hline
FlatVel-B & 0.0210 & 0.0632 & 0.9718  & 0.0233 & 0.0696 & 0.9658  & 0.0193 & 0.0621 & 0.9729 & \textbf{0.0173} & \textbf{0.0584} & \textbf{0.9756} \\ \hline
CurveVel-A & 0.0409 & 0.0944 & 0.8796  & 0.0343 & 0.0798 & 0.9027  & 0.0272 & 0.0725 & 0.9180 & \textbf{0.0260} & \textbf{0.0705} & \textbf{0.9199} \\ \hline
CurveVel-B & 0.1073 & 0.2349 & 0.7527  & 0.0933 & 0.2154 & 0.7808  & 0.0816 & 0.2006 & 0.8053 & \textbf{0.0772} & \textbf{0.1947} & \textbf{0.8134} \\ \hline
FlatFault-A & 0.0096 & 0.0278 & 0.9880  & 0.0106 & 0.0286 & 0.9871  & 0.0075 & 0.0229 & 0.9904 & \textbf{0.0066} & \textbf{0.0208} & \textbf{0.9918} \\ \hline
FlatFault-B & 0.0843 & 0.1497 & 0.7635  & 0.0710 & 0.1321 & 0.8027  & \textbf{0.0636} & \textbf{0.1259} & \textbf{0.8137} & 0.0644 & 0.1269 & 0.8033 \\ \hline
CurveFault-A & 0.0164 & 0.0485 & 0.9712  & 0.0167 & 0.0474 & 0.9712  & 0.0130 & 0.0404 & 0.9771 & \textbf{0.0117} & \textbf{0.0369} & \textbf{0.9801} \\ \hline
CurveFault-B & 0.1444 & 0.2248 & 0.6274  & 0.1245 & 0.2027 & 0.6781  & \textbf{0.1161} & \textbf{0.1954} & \textbf{0.6896} & 0.1169 & 0.1960 & 0.6790 \\ \hline
Style-A & 0.0567 & 0.0947 & 0.8972  & 0.0514 & 0.0868 & 0.9125  & \textbf{0.0480} & \textbf{0.0829} & \textbf{0.9187} & 0.0483 & 0.0831 & 0.9136 \\ \hline
Style-B & 0.0542 & 0.0890 & 0.7646  & 0.0553 & 0.0876 & 0.7567  & \textbf{0.0538} & \textbf{0.0867} & \textbf{0.7600} & 0.0563 & 0.0908 & 0.7429 \\ \thickhline
\end{tabular}
\caption{Quantitative comparison between the results of InversionNet and BigFWIs on OpenFWI-408k in terms of MAE, RMSE and SSIM.}
\label{tab:408k}
\end{table}

\begin{table}[t]
\small
\centering
\setlength{\tabcolsep}{5pt}
\renewcommand{\arraystretch}{1.5}
\begin{tabular}{c|c c|c c|c c|c c}
\thickhline
\multirow{2}{*}{Dataset} & \multicolumn{2}{c|}{InversionNet} & \multicolumn{2}{c|}{BigFWI-B} & \multicolumn{2}{c|}{BigFWI-M} & \multicolumn{2}{c}{BigFWI-L} \\ \cline{2-9} 

& WD$_{Seis}\downarrow$ & WD$_{Vmap}\downarrow$  & WD$_{Seis}\downarrow$ & WD$_{Vmap}\downarrow$ & WD$_{Seis}\downarrow$ & WD$_{Vmap}\downarrow$  & WD$_{Seis}\downarrow$ & WD$_{Vmap}\downarrow$ \\ \thickhline

FlatVel-A    & 7.1659    & 0.0363  & 6.5469     & 0.0319 & \textbf{3.9796}     & 0.0299 & 4.2737     & \textbf{0.0273} \\ \hline
FlatVel-B    & 44.9913   & 0.1059  & 56.9304    & 0.1017 & 37.7425    & 0.0901 & \textbf{34.0136}    & \textbf{0.0875} \\ \hline
CurveVel-A   & 256.9994  & 0.1335  & 215.5561   & 0.0964 & 160.5040   & 0.0851 & \textbf{154.5497}   & \textbf{0.0780} \\ \hline
CurveVel-B   & 732.5136  & 0.3340  & 607.3340   & 0.2661 & 537.0858   & 0.2378 & \textbf{498.6442}   & \textbf{0.2216} \\ \hline
FlatFault-A  & 96.9786   & 0.0358  & 126.8728   & 0.0364 & 90.7467    & 0.0312 & \textbf{80.5870}    & \textbf{0.0293} \\ \hline
FlatFault-B  & 521.3496  & 0.2458  & 440.6822   & 0.1838 & \textbf{345.5444}   & 0.1662 & 362.3163   & \textbf{0.1468} \\ \hline
CurveFault-A & 186.2397  & 0.0507  & 218.2358   & 0.0449 & 166.9954   & 0.0421 & \textbf{152.6674}   & \textbf{0.0373} \\ \hline
CurveFault-B & 1044.2372 & 0.3357  & 891.3471   & 0.2604 & \textbf{773.0494}   & 0.2367 & 773.9282   & \textbf{0.2179} \\ \hline
Style-A      & 125.4812  & 0.1617  & 126.1750   & 0.1426 & \textbf{113.1486}   & 0.1301 & 121.2802   & \textbf{0.1235} \\ \hline
Style-B      & \textbf{96.8782}   & 0.0866  & 169.6936   & 0.0828 & 169.4446   & 0.0787 & 199.0436   & \textbf{0.0784} \\ \thickhline
\end{tabular}
\caption{\color{black}Quantitative comparison between the results of InversionNet and BigFWIs on OpenFWI-408k in terms of Wasserstein Distance.}
\label{tab:408k-wd}
\end{table}

\begin{table}[t]
\centering
\setlength{\tabcolsep}{10pt}
\renewcommand{\arraystretch}{1.5}
\begin{tabular}{c|c c c c|c c c}
\thickhline
\multirow{2}{*}{\makecell{Target\\Dataset}} & \multicolumn{4}{c|}{InversionNet} & \multicolumn{3}{c}{BigFWI} \\ \cline{2-8} 
& Source & MAE$\downarrow$ & RMSE$\downarrow$ & SSIM$\uparrow$  & MAE$\downarrow$ & RMSE$\downarrow$ & SSIM$\uparrow$ \\ \thickhline
FlatVel-A & FlatVel-B & 0.0207 & 0.0381 & 0.9723 & \textbf{0.0137} & \textbf{0.0285} & \textbf{0.9795} \\ \hline
FlatVel-B & CurveVel-B & 0.1076 & 0.2331 & 0.7797 & \textbf{0.0820} & \textbf{0.1970} & \textbf{0.8345} \\ \hline
CurveVel-A & CurveVel-B & 0.0833 & 0.1458 & 0.7828 & \textbf{0.0578} & \textbf{0.1114} & \textbf{0.8404} \\ \hline
CurveVel-B & FlatFault-B & 0.4267 & 0.5649 & 0.4234 & \textbf{0.2543} & \textbf{0.4042} & \textbf{0.5373} \\ \hline
FlatFault-A & CurveFault-A & 0.0394 & 0.0979 & 0.9224 & \textbf{0.0211} & \textbf{0.0626} & \textbf{0.9618} \\ \hline
FlatFault-B & CurveFault-B & 0.1213 & 0.1895 & 0.6677 & \textbf{0.0998} & \textbf{0.1630} & \textbf{0.7343} \\ \hline
CurveFault-A & FlatFault-B & 0.0834 & 0.1537 & 0.8364 & \textbf{0.0398} & \textbf{0.0955} & \textbf{0.9198} \\ \hline
CurveFault-B & FlatFault-B & 0.1898 & 0.2840 & 0.5369 & \textbf{0.1638} & \textbf{0.2528} & \textbf{0.5905} \\ \hline
Style-A & Style-B & 0.1195 & 0.1655 & 0.7653 & \textbf{0.0948} & \textbf{0.1372} & \textbf{0.8049} \\ \hline
Style-B & Style-A & 0.0858 & 0.1226 & 0.6817 & \textbf{0.0801} & \textbf{0.1142} & \textbf{0.6871} \\ \thickhline
\end{tabular}
\caption{Quantitative comparison between the generalization results of InversionNet and BigFWIs (leave-one-out) in terms of MAE, RMSE and SSIM.}
\label{tab:gen}
\end{table}

\begin{table}[t]
\centering
\setlength{\tabcolsep}{10pt}
\renewcommand{\arraystretch}{1.5}
\begin{tabular}{c|c c c|c c}
\thickhline
\multirow{2}{*}{\makecell{Target\\Dataset}} & \multicolumn{3}{c|}{InversionNet} & \multicolumn{2}{c}{BigFWI} \\ \cline{2-6} 
& Source & WD$_{Seis}\downarrow$ & WD$_{Vmap}\downarrow$ & WD$_{Seis}\downarrow$ & WD$_{Vmap}\downarrow$ \\ \thickhline
FlatVel-A & FlatVel-B & \textbf{31.5622} & 0.0912 & 32.2925 & \textbf{0.0616}\\ \hline
FlatVel-B & CurveVel-B & 471.8385 & 0.3892 & \textbf{364.7918} & \textbf{0.3283} \\ \hline
CurveVel-A & CurveVel-B & 631.7191 & 0.2903 & \textbf{472.7896} & \textbf{0.1658} \\ \hline
CurveVel-B & FlatFault-B & 5445.9127 & 1.8209 & \textbf{2597.8299} & \textbf{0.7624} \\ \hline
FlatFault-A & CurveFault-A & 631.0078 & 0.1113 & \textbf{454.4401} & \textbf{0.0801} \\ \hline
FlatFault-B & CurveFault-B & 863.8857 & 0.3363 & \textbf{795.1951} & \textbf{0.2664} \\ \hline
CurveFault-A & FlatFault-B & 1419.7535 & 0.3397 & \textbf{671.2540} & \textbf{0.1189} \\ \hline
CurveFault-B & FlatFault-B & 1703.5499 & 0.4681 & \textbf{1307.8235} & \textbf{0.3566} \\ \hline
Style-A & Style-B & 317.7699 & 0.5207 & \textbf{285.1514} & \textbf{0.3141} \\ \hline
Style-B & Style-A & \textbf{284.4557} & 0.2024 & 406.7438 & \textbf{0.1798} \\ \thickhline
\end{tabular}
\caption{\color{black}Quantitative comparison between the generalization results of InversionNet and BigFWIs (leave-one-out) in terms of Wasserstein Distance.}
\label{tab:gen-wd}
\end{table}

\begin{table}[t]
\small
\centering
\setlength{\tabcolsep}{5pt}
\renewcommand{\arraystretch}{1.5}
\begin{tabular}{c|c c c|c c c|c c c|c c c}
\thickhline
\multirow{2}{*}{Dataset} & \multicolumn{3}{c|}{InversionNet-SA} & \multicolumn{3}{c|}{InversionNet-SB} & \multicolumn{3}{c|}{BigFWI-M} & \multicolumn{3}{c}{BigFWI-L} \\ \cline{2-13} 
& MAE$\downarrow$ & RMSE$\downarrow$ & SSIM$\uparrow$  & MAE$\downarrow$ & RMSE$\downarrow$ & SSIM$\uparrow$ & MAE$\downarrow$ & RMSE$\downarrow$ & SSIM$\uparrow$ & MAE$\downarrow$ & RMSE$\downarrow$ & SSIM$\uparrow$ \\ \thickhline

\makecell{Marmousi\\(smooth)}  & 0.1410 & 0.1996 & 0.7408 & 0.1456 & 0.1988 & 0.5886 & \textbf{0.0792} & \textbf{0.1164} & \textbf{0.8356} & 0.0823 & 0.1244 & 0.7808 \\ \hline
\makecell{Marmousi\\(original)} & 0.1783 & 0.2505 & 0.4749 & 0.2161 & 0.2942 & 0.3798 & \textbf{0.1492} & \textbf{0.2323} &\textbf{0.4936} & 0.1549 & 0.2419 & 0.4806 \\ \hline
\makecell{Overthrust\\(smooth)}  & 0.1213 & 0.1719 & 0.7511 & 0.1062 & 0.1398 & 0.6898 & \textbf{0.0722} & \textbf{0.1000} & \textbf{0.7599} & 0.0760 & 0.1001 & 0.7491 \\ \hline
\makecell{Overthrust\\(original)}  & 0.2052 & 0.2742 & 0.4177 & 0.1819 & 0.2414 & 0.4252 & \textbf{0.1549} & \textbf{0.2040} & \textbf{0.4608} & 0.1775 & 0.2297 & 0.3938 \\ \hline

\end{tabular}
\caption{Quantitative comparison betwen the generalization results of InversionNet and BigFWIs on Marmousi and Overthrust in terms of MAE, RMSE and SSIM.}
\label{tab:gen_marm_over}
\end{table}

\begin{table}[t]
\small
\centering
\setlength{\tabcolsep}{5pt}
\renewcommand{\arraystretch}{1.5}
\begin{tabular}{c|c c|c c|c c|c c}
\thickhline
\multirow{2}{*}{Dataset} & \multicolumn{2}{c|}{InversionNet-SA} & \multicolumn{2}{c|}{InversionNet-SB} & \multicolumn{2}{c|}{BigFWI-M} & \multicolumn{2}{c}{BigFWI-L} \\ \cline{2-9} 
& WD$_{Seis}\downarrow$ & WD$_{Vmap}\downarrow$ & WD$_{Seis}\downarrow$ & WD$_{Vmap}\downarrow$ & WD$_{Seis}\downarrow$ & WD$_{Vmap}\downarrow$ & WD$_{Seis}\downarrow$ & WD$_{Vmap}\downarrow$ \\ \thickhline

\makecell{Marmousi\\(smooth)}  & 1140.5796 & 0.6277 & \textbf{1092.4845} & 0.4691 & 1104.5630 & \textbf{0.2706} & 1125.5029 & 0.2882 \\ \hline
\makecell{Marmousi\\(original)} & 2160.1560 & 0.6764 & 3215.7143 & 0.5241 & 1460.0678 & 0.2840 & \textbf{1293.1754} & \textbf{0.2558} \\ \hline
\makecell{Overthrust\\(smooth)} & 2203.9591 & 0.6695 & 1059.8792 & 0.4295 & 1045.3390 & 0.1136 & \textbf{976.3700} & \textbf{0.0752} \\ \hline
\makecell{Overthrust\\(original)} & 2794.4750 & 0.7481 & \textbf{1376.2766} & 0.4220 & 1805.2714 & \textbf{0.0981} & 1697.1583 & 0.1567 \\ \hline

\end{tabular}
\caption{\color{black}Quantitative comparison between the generalization results of InversionNet and BigFWIs on Marmousi and Overthrust in terms of Wasserstein Distance.}
\label{tab:gen_marm_over-wd}
\end{table}

\begin{table}[h]
\centering
\setlength{\tabcolsep}{3pt}
\renewcommand{\arraystretch}{2.0}
\small
\begin{tabular}{c |c c|c c|c c|c c}
\thickhline
\multirow{2}{*}{Dataset} & \multicolumn{2}{c|}{InversionNet-SA} & \multicolumn{2}{c|}{InversionNet-SB} & \multicolumn{2}{c|}{BigFWI-M} & \multicolumn{2}{c}{BigFWI-L} \\ \cline{2-9} 
& \makecell{RMS$_{RTM}\downarrow$\\($\times10^{-3})$} & \makecell{2-norm$_{RTM}\downarrow$\\($\times10^{-1})$} & \makecell{RMS$_{RTM}\downarrow$\\($\times10^{-3})$} & \makecell{2-norm$_{RTM}\downarrow$\\($\times10^{-1})$} & \makecell{RMS$_{RTM}\downarrow$\\($\times10^{-3})$} & \makecell{2-norm$_{RTM}\downarrow$\\($\times10^{-1})$} & \makecell{RMS$_{RTM}\downarrow$\\($\times10^{-3})$} & \makecell{2-norm$_{RTM}\downarrow$\\($\times10^{-1})$} \\ \thickhline

\makecell{Marmousi\\(smooth)}  & 0.8396 & 0.5877 & 1.3680 & 0.9576 & \textbf{0.5962} & \textbf{0.4174} & 0.7789 & 0.5452 \\ \hline
\makecell{Marmousi\\(original)} & 2.1476 & 1.5033 & 2.3882 & 1.6717 & \textbf{2.0772} & \textbf{1.4540} & 2.1852 & 1.5296 \\ \hline
\makecell{Overthrust\\(smooth)} & \textbf{0.9596} & \textbf{0.6717} & 1.1037 & 0.7726 & 1.0319 & 0.7223 & 1.0512 & 0.7359 \\ \hline
\makecell{Overthrust\\(original)} & 2.7633 & 1.9343 & 2.7022 & 1.8915 & \textbf{2.5804} & \textbf{1.8063} & 2.7460 & 1.9222 \\ \thickhline

\end{tabular}
\caption{\color{black}Quantitative comparison between the generalization results of InversionNet and BigFWIs on Marmousi and Overthrust in terms of Wasserstein Distance.}
\label{tab:gen_marm_over-rtm}
\end{table}

\begin{figure}
    \centering
    \includegraphics[width=\textwidth]{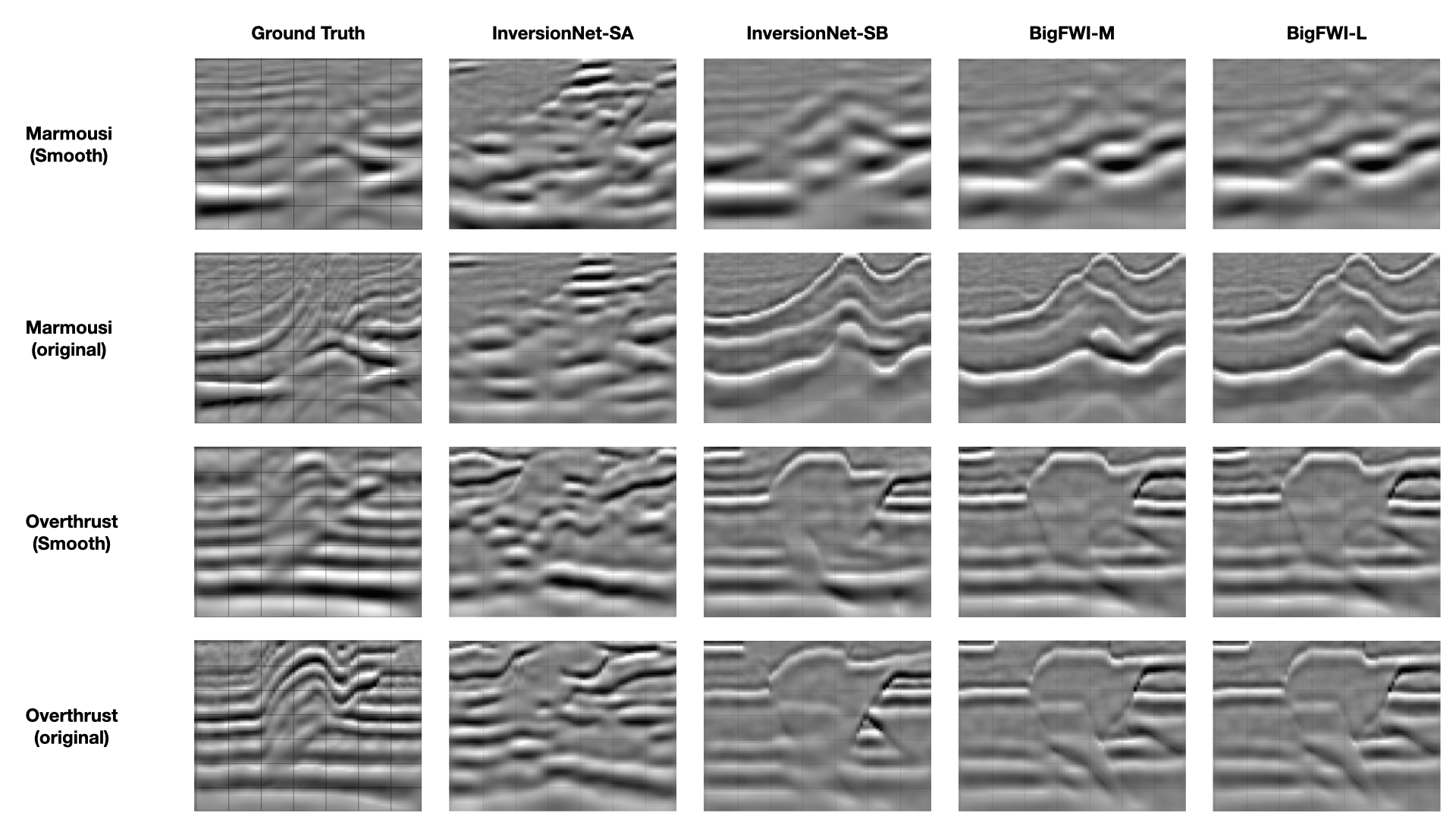}
    \caption{\color{black}Zero-offset least square reverse time migration (LSRTM) images using different velocity models for comparison.}
    \label{fig:rtm}
\end{figure}

\begin{figure}
    \centering
    \includegraphics[width=\textwidth]{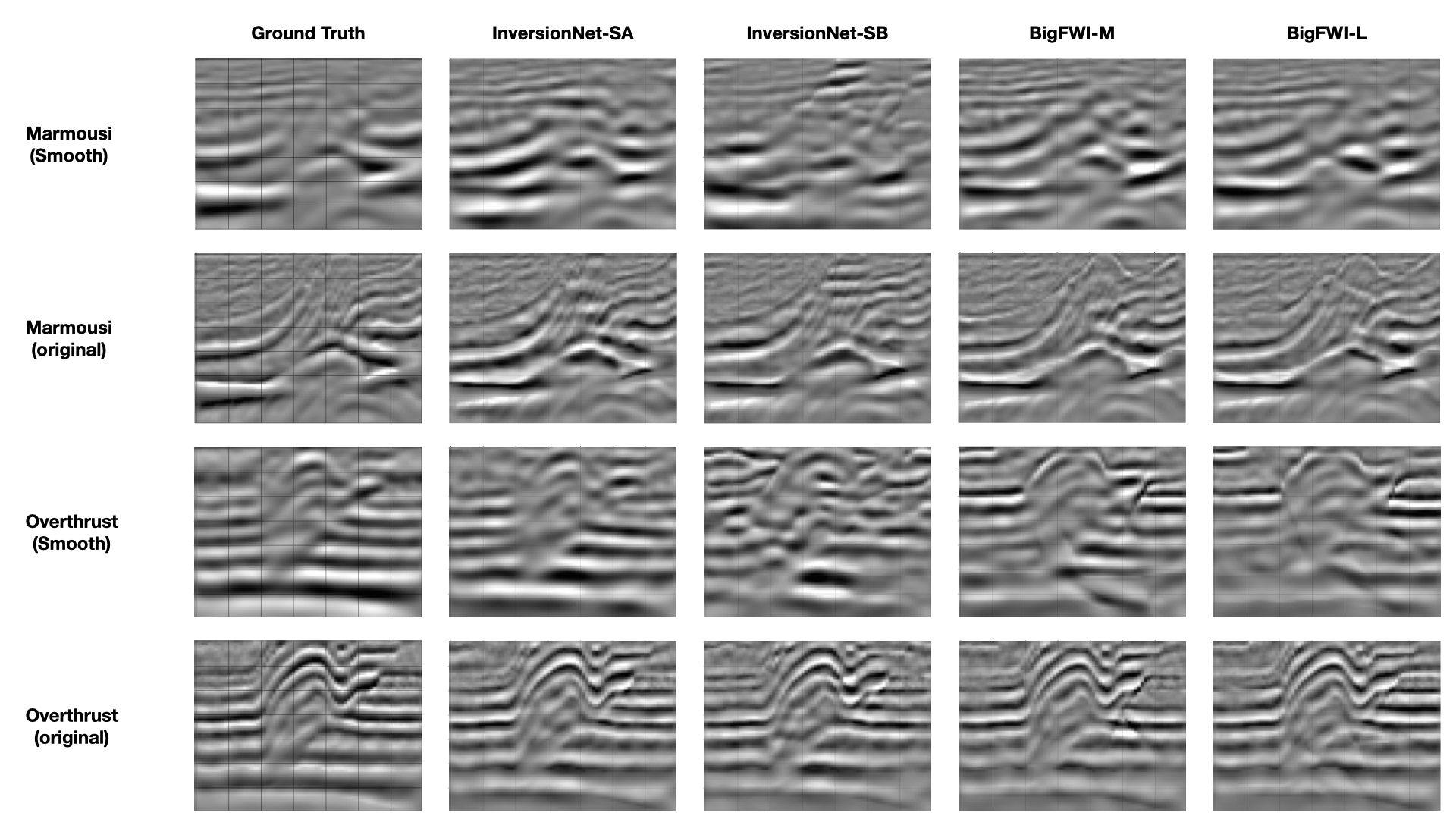}
    \caption{\color{black}Zero-offset least square reverse time migration (LSRTM) image differences to the ground truth images using different velocity models for comparison.}
    \label{fig:rtm_diff}
\end{figure}
\clearpage
\bibliography{supplement}